\documentclass[10pt,twocolumn,letterpaper]{article}

\usepackage[pagenumbers]{wacv} 

\usepackage{times}
\usepackage{epsfig}
\usepackage{graphicx}
\usepackage{amsmath}
\usepackage{amssymb}
\usepackage{booktabs} 
\usepackage{multirow}

\definecolor{wacvblue}{rgb}{0.21,0.49,0.74}
\usepackage[pagebackref,breaklinks,colorlinks,allcolors=wacvblue]{hyperref}

\AtBeginDocument{%
  \crefname{figure}{Figure}{Figures}%
  \Crefname{figure}{Figure}{Figures}%
  \crefname{table}{Table}{Tables}%
  \Crefname{table}{Table}{Tables}%
  \crefname{section}{Section}{Sections}%
  \Crefname{section}{Section}{Sections}%
}

\def\wacvPaperID{} 
\def\confName{WACV}
\def\confYear{2026}

\title{SPoRC-VIST: A Benchmark for Evaluating Generative Natural Narrative in Vision-Language Models}

\author{Yunlin Zeng\\
Georgia Institute of Technology\\
{\tt\small yzeng@gatech.edu}
}

\begin{document}
\maketitle

\begin{abstract}
Vision-Language Models (VLMs) have achieved remarkable success in descriptive tasks such as image captioning and visual question answering (VQA). However, their ability to generate engaging, long-form narratives---specifically multi-speaker podcast dialogues---remains under-explored and difficult to evaluate. Standard metrics like BLEU and ROUGE fail to capture the nuances of conversational naturalness, personality, and narrative flow, often rewarding safe, repetitive outputs over engaging storytelling. In this work, we present a novel pipeline for end-to-end visual podcast generation, and fine-tune a Qwen3-VL-32B model on a curated dataset of 4,000 image-dialogue pairs. Crucially, we use a synthetic-to-real training strategy: we train on high-quality podcast dialogues from the Structured Podcast Research Corpus (SPoRC) paired with synthetically generated imagery, and evaluate on real-world photo sequences from the Visual Storytelling Dataset (VIST). This rigorous setup tests the model's ability to generalize from synthetic training data to real-world visual domains. We propose a comprehensive evaluation framework that moves beyond textual overlap, and use AI-as-a-judge (Gemini 3 Pro, Claude Opus 4.5, GPT 5.2) and novel style metrics (average turn length, speaker switch rate) to assess quality. Our experiments demonstrate that our fine-tuned 32B model significantly outperforms a 235B base model in conversational naturalness ($>$80\% win rate) and narrative depth (+50\% turn length), while maintaining identical visual grounding capabilities (CLIPScore: 20.39).
\end{abstract}

\section{Introduction}

The field of Computer Vision has rapidly evolved from passive perception (classification, detection) to active generation. Modern Vision-Language Models (VLMs) are capable of processing complex visual inputs and generating detailed textual descriptions. However, a significant gap remains between \textit{description} and \textit{storytelling}. While state-of-the-art models can accurately identify ``a white bus in a forest,'' they often struggle to weave that visual cue into an engaging, multi-turn conversation that exhibits personality, humor, and natural flow.

This limitation is partly due to training data---most VLMs are trained on caption-heavy datasets like LAION~\cite{schuhmann2022laion} or COCO~\cite{lin2014coco}, which prioritize factual brevity---and partly due to the lack of appropriate evaluation metrics for narrative quality. Standard n-gram metrics (BLEU~\cite{papineni2002bleu}, ROUGE~\cite{lin2004rouge}) penalize creativity and linguistic diversity, effectively encouraging models to produce safe, repetitive, and robotic outputs. As Generative AI moves into creative domains, assessing the ``quality'' of a generated narrative requires new frameworks that account for hallucinations of personality, conversational dynamics, and prosodic structure.

\begin{figure*}[t]
\centering
\scriptsize
\textbf{Row 1: Wedding Party (Story ID: 50242)}\\[2pt]
\begin{tabular}{@{}c@{\hspace{3pt}}c@{\hspace{3pt}}c@{\hspace{3pt}}c@{\hspace{3pt}}c@{}}
\includegraphics[width=0.185\linewidth]{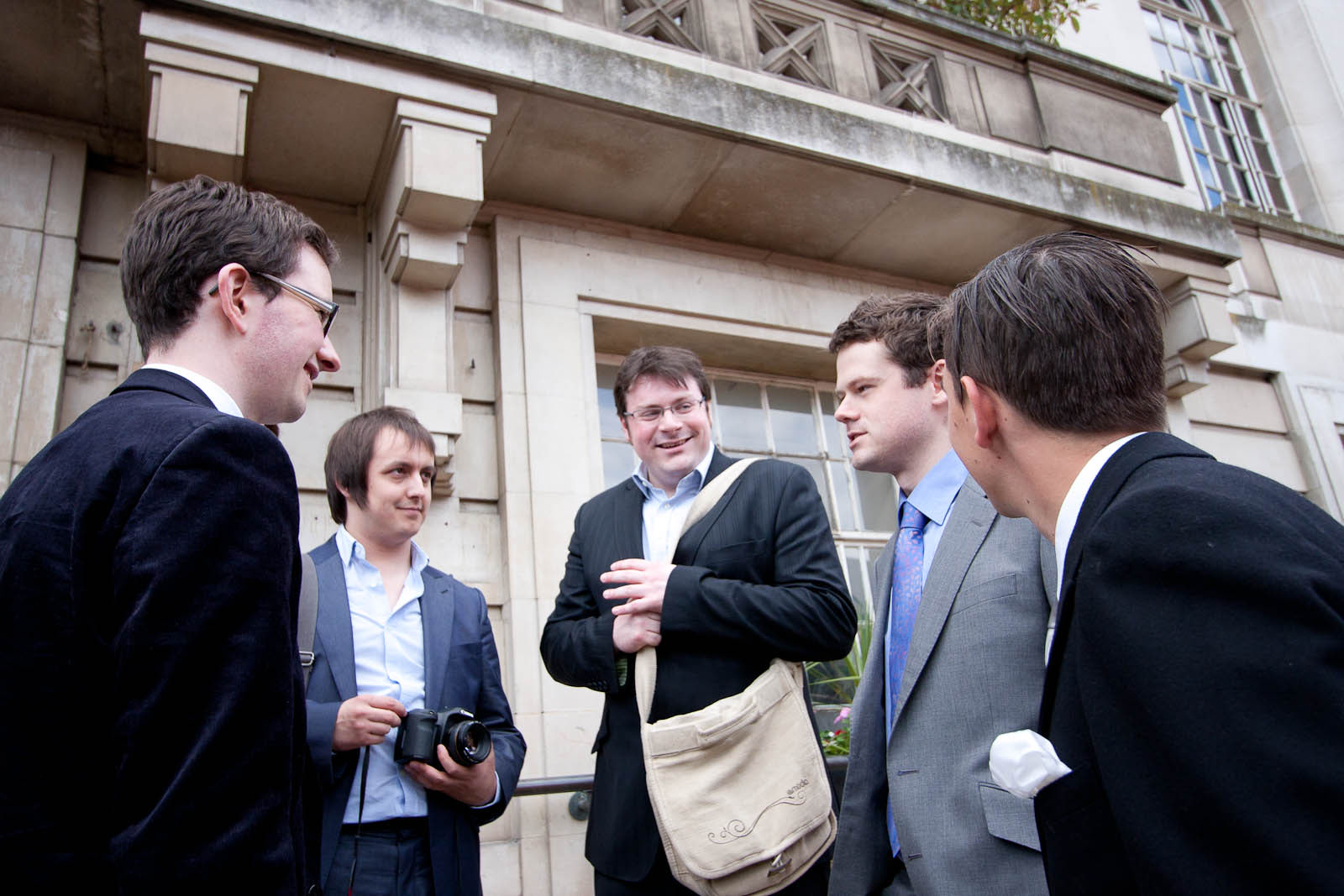} &
\includegraphics[width=0.185\linewidth]{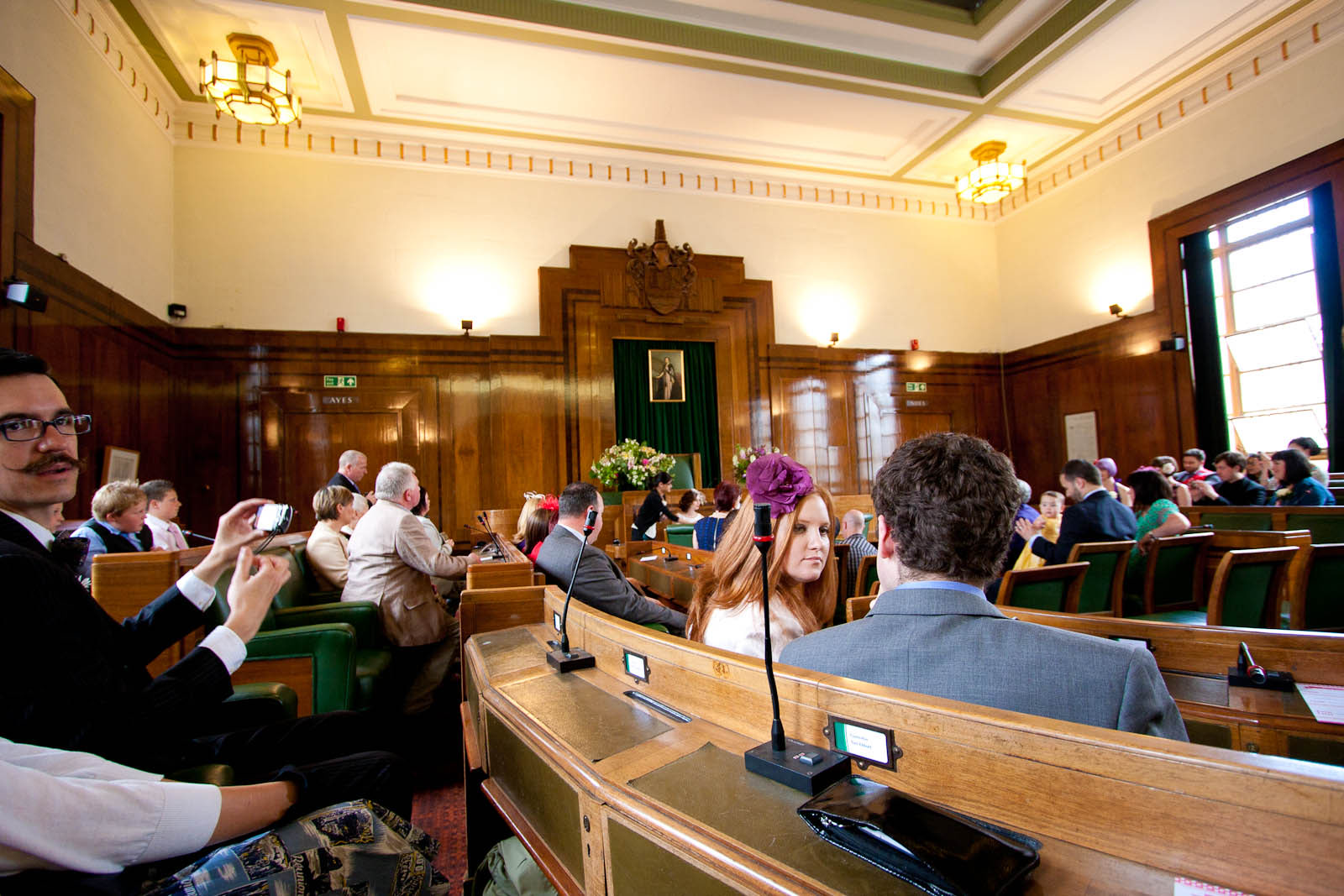} &
\includegraphics[width=0.185\linewidth]{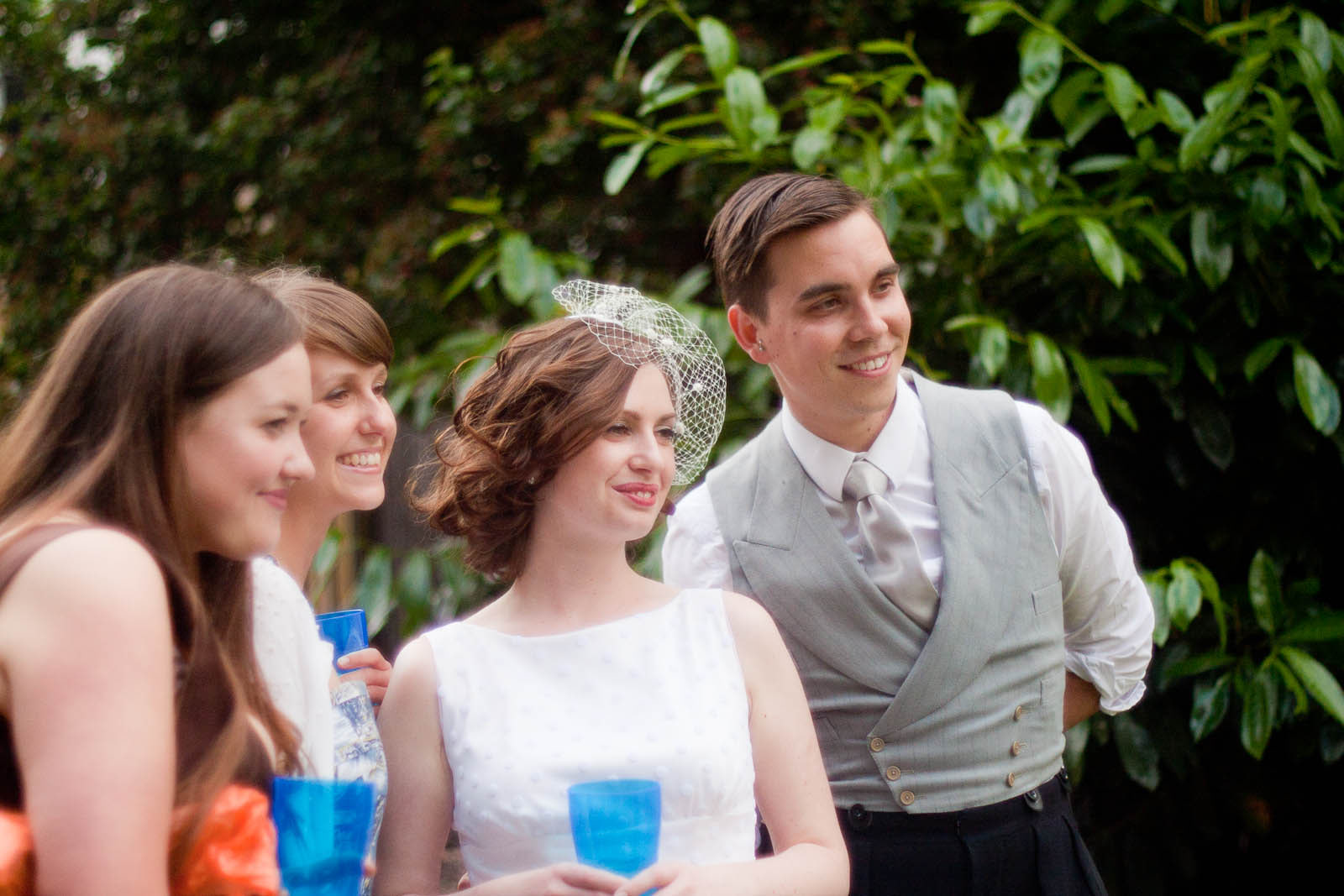} &
\includegraphics[width=0.185\linewidth]{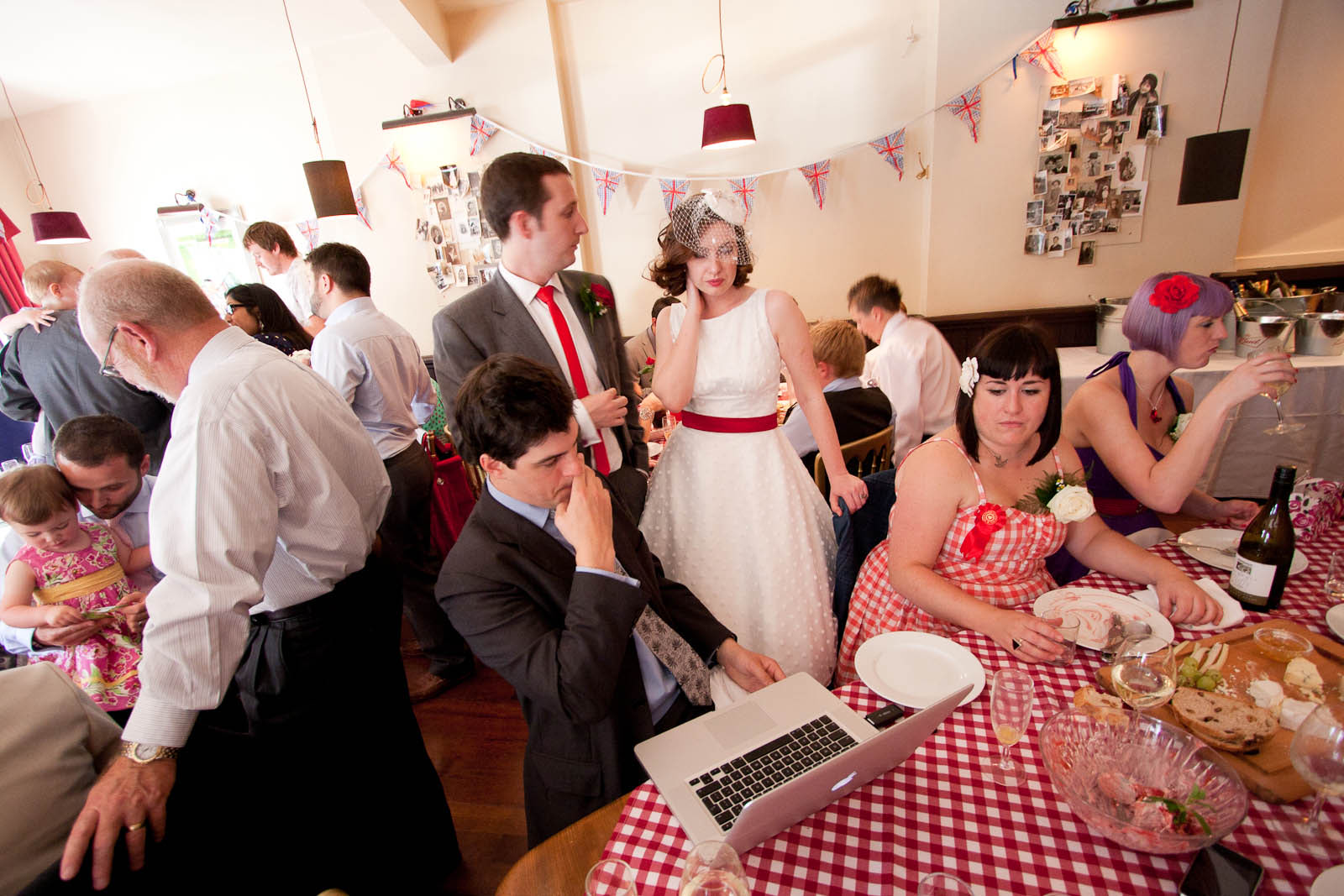} &
\includegraphics[width=0.185\linewidth]{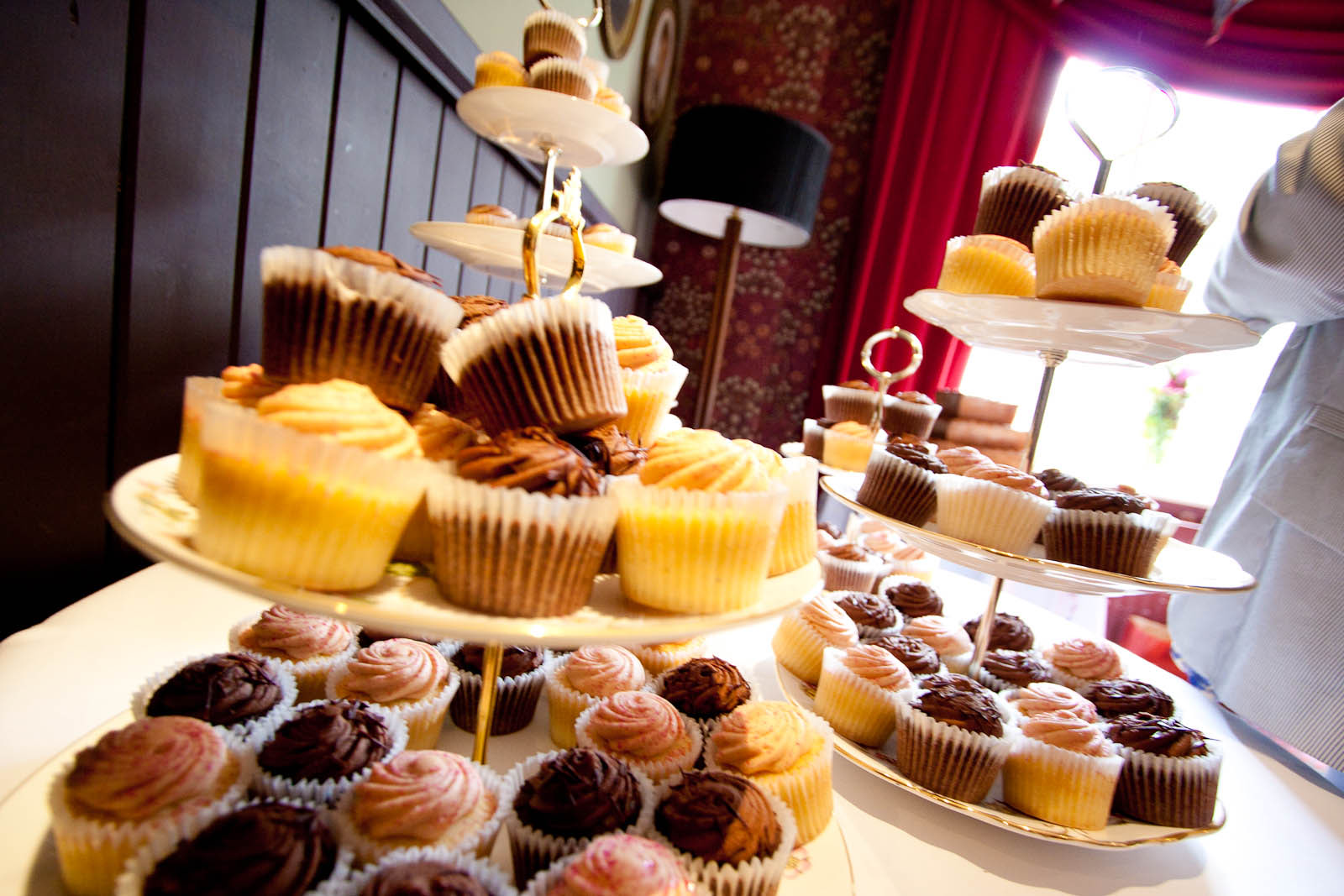} \\[-1pt]
\parbox{0.185\linewidth}{\centering (1) ``I went down to the wedding party.''} &
\parbox{0.185\linewidth}{\centering (2) ``There were tons of people there.''} &
\parbox{0.185\linewidth}{\centering (3) ``We were having some drinks outside.''} &
\parbox{0.185\linewidth}{\centering (4) ``There was a lot of food.''} &
\parbox{0.185\linewidth}{\centering (5) ``I ate a lot of cupcakes.''}
\end{tabular}\\[8pt]
\textbf{Row 2: Motorcycle \& Marathon (Story ID: 47306)}\\[2pt]
\begin{tabular}{@{}c@{\hspace{3pt}}c@{\hspace{3pt}}c@{\hspace{3pt}}c@{\hspace{3pt}}c@{}}
\includegraphics[width=0.185\linewidth]{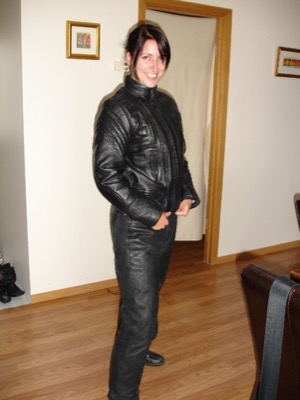} &
\includegraphics[width=0.185\linewidth]{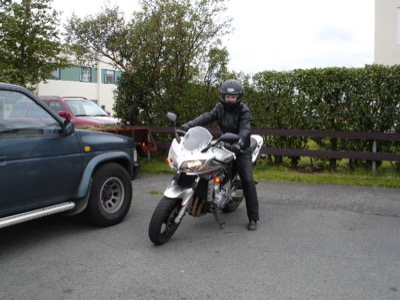} &
\includegraphics[width=0.185\linewidth]{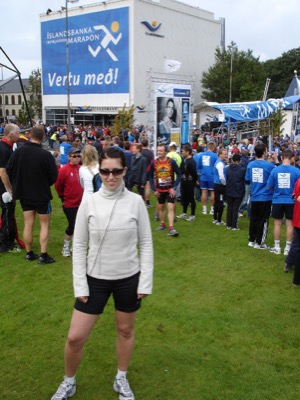} &
\includegraphics[width=0.185\linewidth]{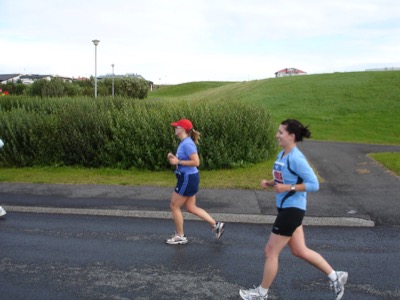} &
\includegraphics[width=0.185\linewidth]{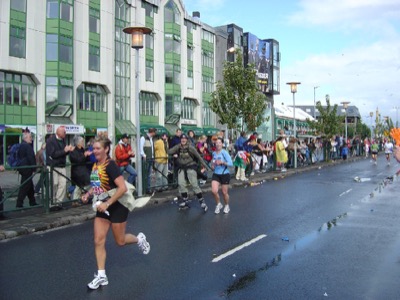} \\[-1pt]
\parbox{0.185\linewidth}{\centering (1) ``This is me on my new motorcycle.''} &
\parbox{0.185\linewidth}{\centering (2) ``I love riding around town.''} &
\parbox{0.185\linewidth}{\centering (3) ``I also ran a marathon.''} &
\parbox{0.185\linewidth}{\centering (4) ``It was raining but I kept going.''} &
\parbox{0.185\linewidth}{\centering (5) ``I finished with a smile.''}
\end{tabular}
\caption{Two evaluation sequences from the VIST dataset. Each row shows five real-world photos with their original single-sentence VIST captions. \textit{Row 1} depicts a wedding celebration; \textit{Row 2} shows motorcycle riding and marathon running. Our goal is to transform these brief, factual captions into rich, multi-turn podcast dialogues. Note: The model is trained on \textit{synthetic} images (Stable Diffusion) but evaluated on these \textit{real} photos to test domain generalization.}
\label{fig:teaser}
\end{figure*}

In this paper, we address the challenge of visual podcast generation: transforming a sequence of images into a coherent, entertaining podcast script between two distinct hosts. \cref{fig:teaser} illustrates a typical input from the Visual Storytelling (VIST) dataset~\cite{huang2016visual} (five images, each with a simple one-sentence caption), which we aim to transform into rich multi-turn dialogues.

We introduce the SPoRC-VIST Benchmark, a framework that uses the abundance of high-quality text data by pairing it with synthetic visuals for training, while testing on real-world photographic sequences. Our contributions are threefold: (1) We curate a dataset of 4,000 visual-dialogue pairs and fine-tune a parameter-efficient Qwen3-VL-32B model using LoRA to perform style transfer from ``captioner'' to ``podcaster.'' (2) We propose a new set of style-aware metrics (turn length, switch rate) and an AI-as-a-Judge protocol to evaluate ``Hallucination of Personality'' and conversational naturalness. (3) We demonstrate that a smaller, fine-tuned model (32B) can outperform a massive base model (235B) in narrative quality without degrading visual grounding performance, and validate the effectiveness of our synthetic-to-real generalization strategy. Code to reproduce data generation and model training is available at \url{https://github.com/Yunlin-Zeng/visual-podcast-VLM}.

\section{Related Work}

\subsection{Visual Storytelling}
The task of generating narratives from image sequences was formalized by the VIST dataset~\cite{huang2016visual}. While VIST established sequential visual storytelling, its annotations consist of a sequence of descriptive single sentences per image. Other approaches have focused on generating a single, paragraph-length story for a set of images. Our work diverges from these fields by structuring the narrative as a multi-speaker dialogue, a significantly more complex task that requires modeling conversational flow, personality, and inter-speaker dynamics, which are not the primary focus of existing visual storytelling benchmarks.

\subsection{Vision-Language Models}
The architecture of VLMs has evolved from early dual-encoder frameworks like CLIP~\cite{radford2021learning} to sophisticated autoregressive models. Deep Fusion approaches such as Flamingo~\cite{alayrac2022flamingo} and OpenFlamingo~\cite{awadalla2023openflamingo} utilized gated cross-attention layers to inject visual information into frozen LLMs, hence enabling strong few-shot performance. Subsequent instruction-tuned models like LLaVA~\cite{liu2023llava} and InstructBLIP~\cite{dai2023instructblip} adopted an early-fusion strategy, projecting image patches into the language model's embedding space as ``visual tokens,'' allowing for more fine-grained reasoning. Proprietary models such as GPT-4V~\cite{openai2023gpt4v} and Gemini Pro Vision~\cite{team2023gemini} have further pushed the boundaries of multi-modal understanding, though their closed nature limits architectural customization.

Most recently, the Qwen-VL series~\cite{bai2023qwenvl,wang2024qwen2vl,bai2025qwen} has established a new state-of-the-art for open-weights models, with Qwen3-VL-235B outperforming GPT-5 on multiple visual reasoning benchmarks. Qwen3-VL introduces dynamic resolution capabilities, allowing the model to process images at their native aspect ratios without aggressive resizing or cropping. This feature is particularly critical for narrative tasks, where small visual details (e.g., facial expressions, background text on a menu) can serve as pivotal plot points. Our work leverages the Qwen3-VL architecture for this reason, utilizing its robust visual encoder to ground open-ended storytelling in precise visual evidence.

\subsection{Generative Evaluation}
Evaluating open-ended text generation is notoriously difficult. Traditional metrics correlate poorly with human judgment for creative tasks. Recent work has proposed ``LLM-as-a-Judge'' frameworks~\cite{zheng2023judging} by using frontier models (e.g., GPT-5~\cite{openai2025gpt5}) to evaluate coherence, hallucination, and style. We adopt this strategy to assess the specific ``podcast persona'' of our models, and argue that style metrics are more relevant than reference-based overlap for this task.

\section{The SPoRC-VIST Benchmark}

To train a model capable of visual banter without relying on scarce paired video-transcript data, we devised a synthetic-to-real data construction strategy. This allows us to leverage the massive abundance of high-quality podcast audio while ensuring the model learns to ground its dialogue in visual concepts.

\subsection{Data Sources}
We utilize the Structured Podcast Research Corpus (SPoRC)~\cite{litterer2024mapping} as our primary source for high-quality dialogue. SPoRC contains transcripts and metadata for over 1.1 million podcast episodes. We filtered this corpus to identify high-quality interaction pairs (Host/Guest) with engaging topics, and excluded monologues or low-quality automated transcripts. This yielded dialogues rich in banter, interruptions, and natural speech patterns, providing the textual foundation for our narrative style transfer.

For the visual component, we generated synthetic imagery to match the SPoRC transcripts. Since the original podcast data lacks paired visuals, we used Stable Diffusion 3.5~\cite{rombach2022high} to create high-fidelity image sequences corresponding to the visual descriptions found within the transcripts. This allowed us to construct a massive, perfectly aligned training set where the visual content explicitly matches the spoken narrative. For evaluation, we use the VIST dataset~\cite{huang2016visual}, which contains sequences of real-world photos from Flickr albums. This serves as our out-of-distribution test set to verify domain generalization: can a model trained on synthetic images describe real-world photos with the same podcast flair?

\subsection{Why Real Transcripts? A Preliminary Study}

A natural question arises: why use real podcast transcripts rather than generating synthetic dialogues with LLMs? To answer this, we conducted a preliminary experiment comparing real transcripts against synthetic alternatives.

\begin{figure}[t]
\centering
\includegraphics[width=0.48\linewidth]{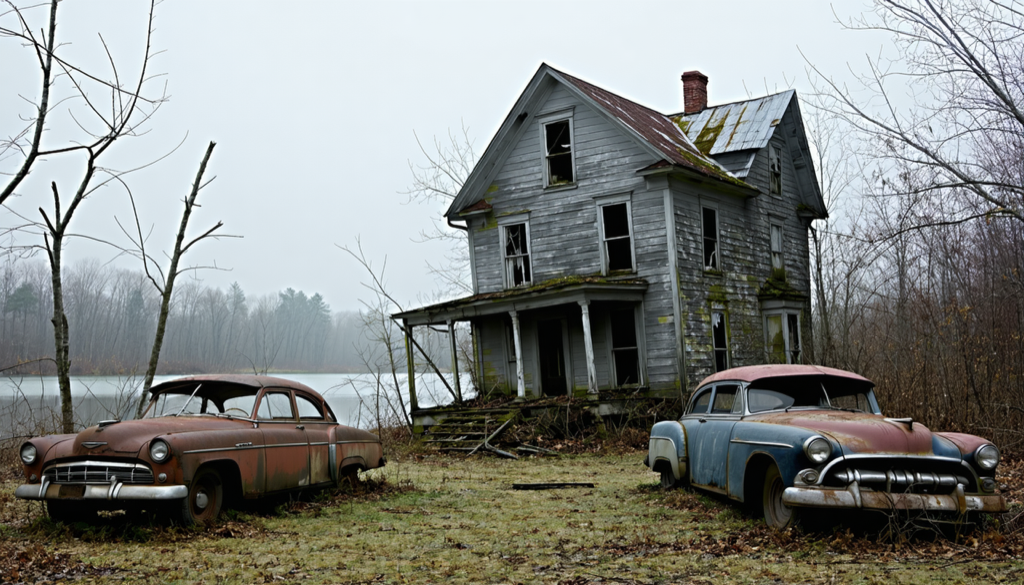}
\includegraphics[width=0.48\linewidth]{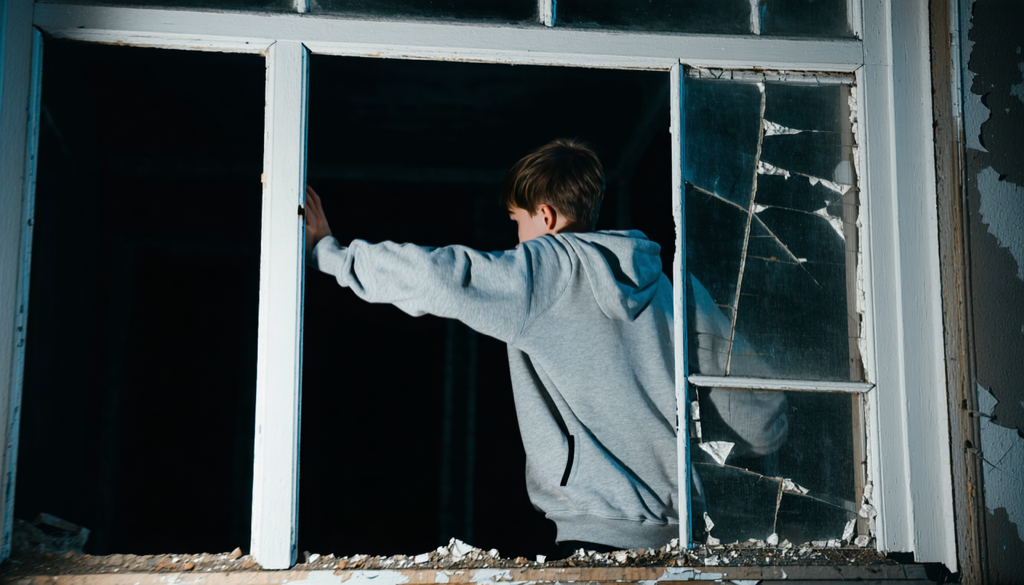}\\[2pt]
\includegraphics[width=0.48\linewidth]{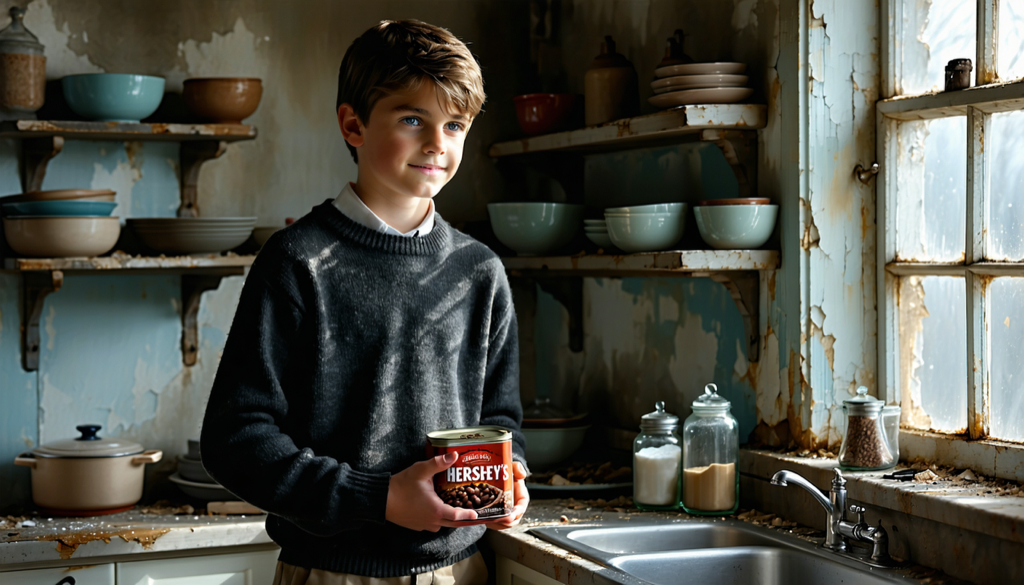}
\includegraphics[width=0.48\linewidth]{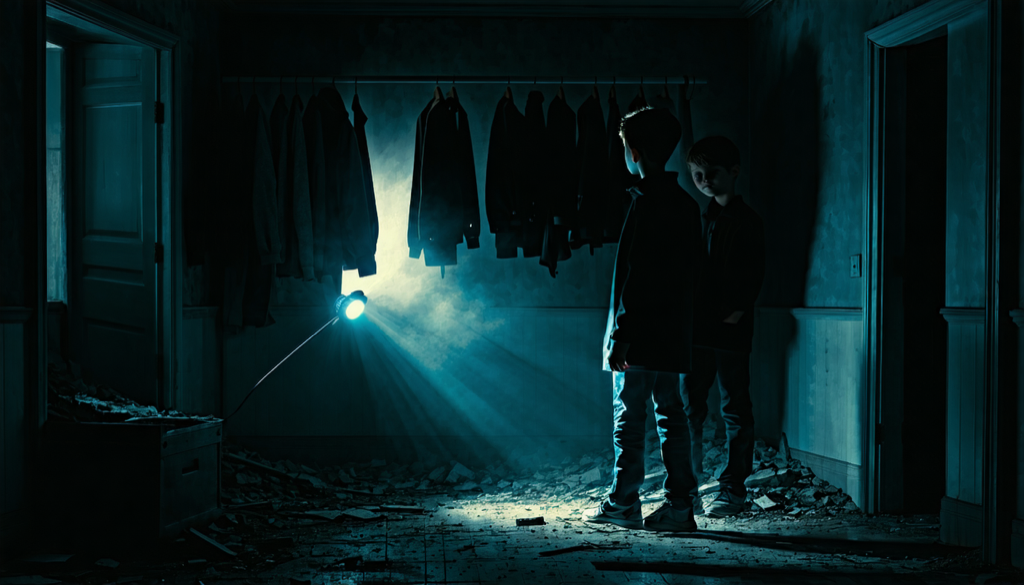}\\[2pt]
\includegraphics[width=0.48\linewidth]{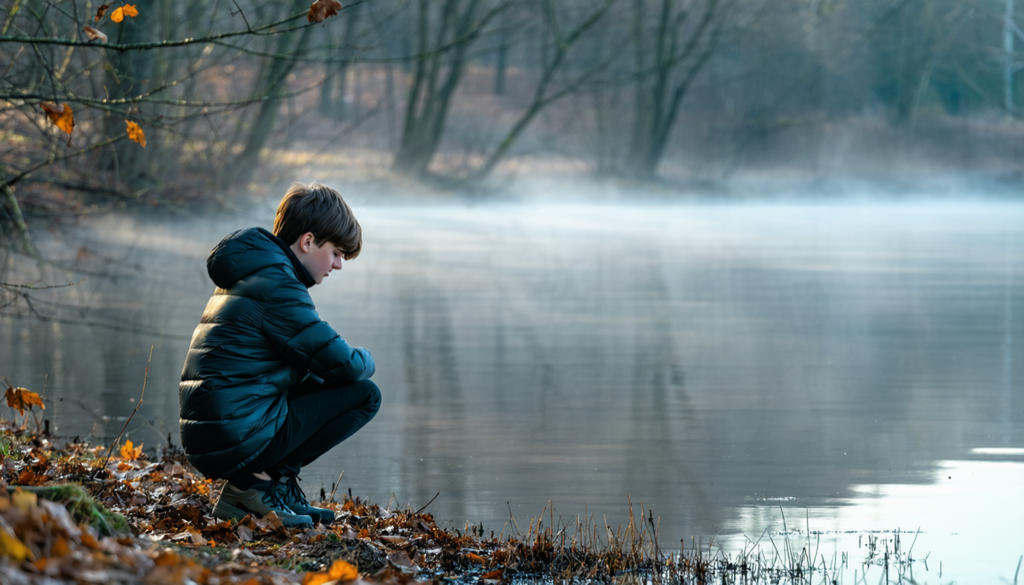}
\caption{Images from our preliminary study. We generated these images from a real podcast excerpt about exploring an abandoned house.\protect\footnotemark{} We then asked AI models to generate transcripts from the images. The original human transcript consistently outperformed AI-generated alternatives, motivating our use of real transcripts as training targets.}
\label{fig:real_vs_synthetic}
\end{figure}

We selected podcast excerpts containing vivid visual descriptions, generated corresponding images with Stable Diffusion 3.5 (\cref{fig:real_vs_synthetic}), and then asked state-of-the-art models (Qwen3-VL-235B and GPT-5.2, chosen for their leading performance on visual reasoning benchmarks) to generate podcast transcripts from these images. We compared the AI-generated transcripts against the original human transcripts using multiple evaluators.

Across all prompt variations, the original human transcripts consistently outperformed AI-generated alternatives in naturalness, personality, and conversational flow. AI-generated dialogues tended to be formulaic (``Welcome to our podcast...'') and lacked the genuine interruptions, personal anecdotes, and emotional reactions that characterize authentic conversation. This finding motivated our decision to use real podcast transcripts as training targets, treating the task as \textit{style transfer} rather than \textit{dialogue generation}.

\footnotetext{Source: \textit{This American Life}, Episode 199 -- ``House on Loon Lake.'' Transcript available at \url{https://www.thisamericanlife.org/199/transcript}}

\subsection{Dataset Construction Pipeline}

\begin{figure*}[t]
\centering
\tiny
\hspace{0.01\textwidth}
\begin{minipage}[t]{0.38\textwidth}
\centering
\scriptsize\textbf{Podcast Excerpt (SPoRC)}\\[2pt]
\hrule\vspace{2pt}
\fontsize{5.5}{6.5}\selectfont\raggedright
\textbf{Host:} Exactly! Now I'm cracking three eggs right into this well. See how the bright yellow yolks are sitting there, perfectly intact? And now I'm taking my fork and I'm just gently beating the eggs, slowly incorporating the flour from the sides.\\[0.5pt]
\textbf{Guest:} The color is amazing---that golden yellow mixing with the white flour. It's almost like watching paint blend.\\[0.5pt]
\textbf{Host:} Right? And now I'm using my hands to bring it all together. The dough is still pretty shaggy at this point---you can see all these little bits of flour still on the board. My hands are completely covered in sticky dough.\\[0.5pt]
\textbf{Guest:} How long do you knead it for?\\[0.5pt]
\textbf{Host:} About 10 minutes. Watch---I'm pushing it away with the heel of my hand, folding it back, turning it a quarter turn. Push, fold, turn. Push, fold, turn. You can see it's getting smoother and more elastic now.\\[0.5pt]
\textbf{Guest:} It's changed so much! It went from that rough, shaggy mess to this smooth, silky ball.\\[0.5pt]
\textbf{Host:} Now I'm wrapping it in this plastic wrap and we'll let it rest for 30 minutes. While we wait, let me show you the pasta machine I'm going to use. It's this silver hand-crank one that clamps right onto the counter here.\\[0.5pt]
\textbf{Guest:} That's a beauty! Is that vintage?\\[0.5pt]
\textbf{Host:} It's my grandmother's actually. Still works perfectly. See these rollers here? We'll run the dough through, starting at the widest setting. But first, let me unwrap this dough---wow, it's so smooth now, feel how soft it is.\\[0.5pt]
\textbf{Guest:} Like silk!\\[0.5pt]
\textbf{Host:} I'm cutting it into four pieces. Now I'm taking this first piece and flattening it with my hands into a rough rectangle shape, about a quarter inch thick. I'm dusting it with flour so it doesn't stick. Now I'm feeding it through the widest setting on the machine.\\[0.5pt]
\textbf{Guest:} Oh wow, it's coming out the other side!\\[0.5pt]
\textbf{Host:} Yeah! And look how it stretches. Now I'm folding it in thirds like a letter, and I'll run it through again. We'll do this a few times. Each time it gets smoother and more uniform. See how it's going from this rough sheet to something really refined?\\[0.5pt]
\textbf{Guest:} The transformation is incredible. It's like watching fabric being woven.\\[0.5pt]
\textbf{Host:} Now I'm moving to the next setting, making it thinner. Watch how long the sheet gets as it comes through. I have to catch it with both hands now. It's getting so delicate I can almost see through it.\\[0.5pt]
\textbf{Guest:} Are you nervous it'll tear?\\[0.5pt]
\textbf{Host:} A little! This is the trickiest part. One more pass... there! Perfect! Now look at this---I'm holding up this long sheet of pasta, and you can see the window behind it. It's almost translucent.\\[0.5pt]
\textbf{Guest:} That's restaurant quality right there!\\[0.5pt]
\textbf{Host:} Thank you! Now I'm switching to the cutting attachment. This one makes fettuccine. Watch---I'm feeding the sheet through and these long ribbons are coming out the bottom. I'm catching them and laying them over this wooden drying rack here.\\[0.5pt]
\textbf{Guest:} They're draped over there like little golden curtains.\\[0.5pt]
\textbf{Host:} Exactly! And that's fresh pasta. From a pile of flour to this in about an hour.
\end{minipage}%
\hfill
\begin{minipage}[t]{0.02\textwidth}
\centering
\vspace{0.8cm}
$\Rightarrow$\\[1.2cm]
\vspace{0.8cm}
$\Rightarrow$\\[1.2cm]
\vspace{0.8cm}
$\Rightarrow$\\[1.2cm]
\vspace{0.8cm}
$\Rightarrow$\\[1.2cm]
\vspace{0.8cm}
$\Rightarrow$
\end{minipage}%
\hfill
\begin{minipage}[t]{0.28\textwidth}
\centering
\scriptsize\textbf{Generated Image Prompts}\\[2pt]
\hrule\vspace{2pt}
\fontsize{5.5}{6.5}\selectfont\raggedright
\it{Scene 1:} A wooden cutting board with three bright yellow egg yolks sitting in a well of white all-purpose flour, with a fork just beginning to beat the eggs, golden yellow starting to swirl into the white flour around the edges.\\[4pt]
\vspace{1.3cm}
\it{Scene 2:} Hands covered in sticky, shaggy pasta dough on a wooden board, with bits of white flour scattered around, the dough still rough and unformed with visible dry patches mixed with wet areas.\\[4pt]
\vspace{1.3cm}
\it{Scene 3:} Hands kneading smooth pasta dough on a wooden board, pushing it away with the heel of the palm in a rhythmic motion, the dough now formed into an elastic, smooth ball that's transformed from its previous rough state.\\[4pt]
\vspace{1.3cm}
\it{Scene 4:} A vintage silver hand-crank pasta machine clamped to a kitchen counter, with a thin sheet of pasta dough being fed through the rollers, the dough stretching and emerging longer on the other side as hands guide it through.\\[4pt]
\vspace{1.3cm}
\it{Scene 5:} A wooden drying rack with long ribbons of fresh fettuccine pasta draped over it like golden curtains, the delicate strands hanging down, nearly translucent with a window visible in the background.
\end{minipage}%
\hfill
\begin{minipage}[t]{0.02\textwidth}
\centering
\vspace{0.8cm}
$\Rightarrow$\\[1.2cm]
\vspace{0.8cm}
$\Rightarrow$\\[1.2cm]
\vspace{0.8cm}
$\Rightarrow$\\[1.2cm]
\vspace{0.8cm}
$\Rightarrow$\\[1.2cm]
\vspace{0.8cm}
$\Rightarrow$
\end{minipage}%
\hfill
\begin{minipage}[t]{0.18\textwidth}
\centering
\scriptsize\textbf{SD 3.5 Output}\\[2pt]
\hrule\vspace{2pt}
\includegraphics[width=1.2\linewidth]{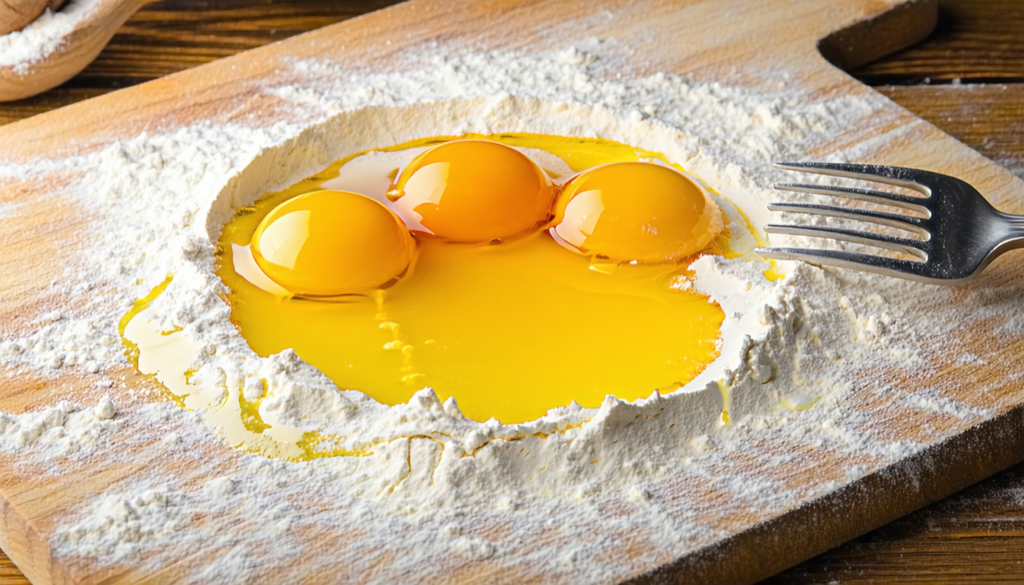}\\[1pt]
\includegraphics[width=1.2\linewidth]{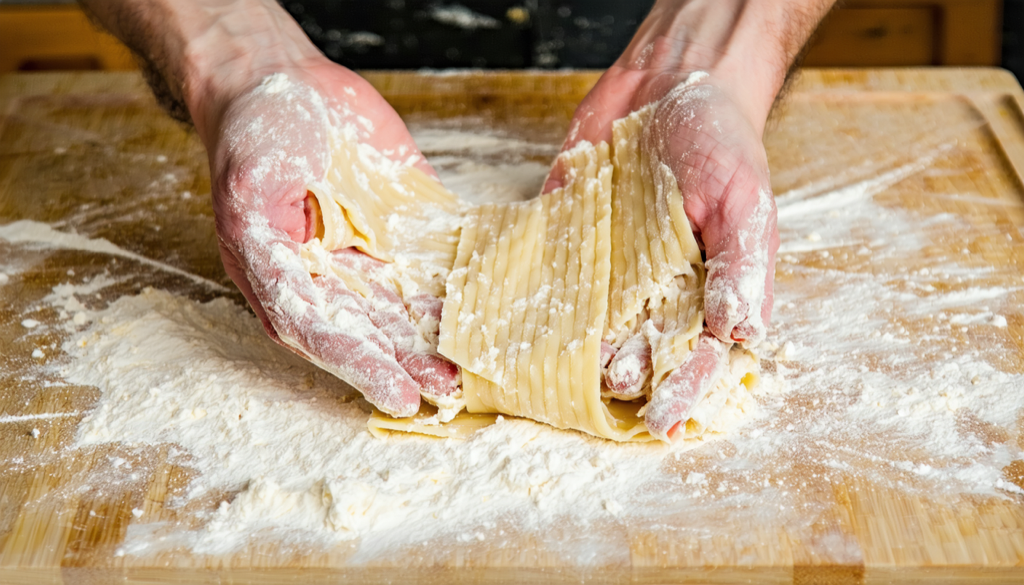}\\[1pt]
\includegraphics[width=1.2\linewidth]{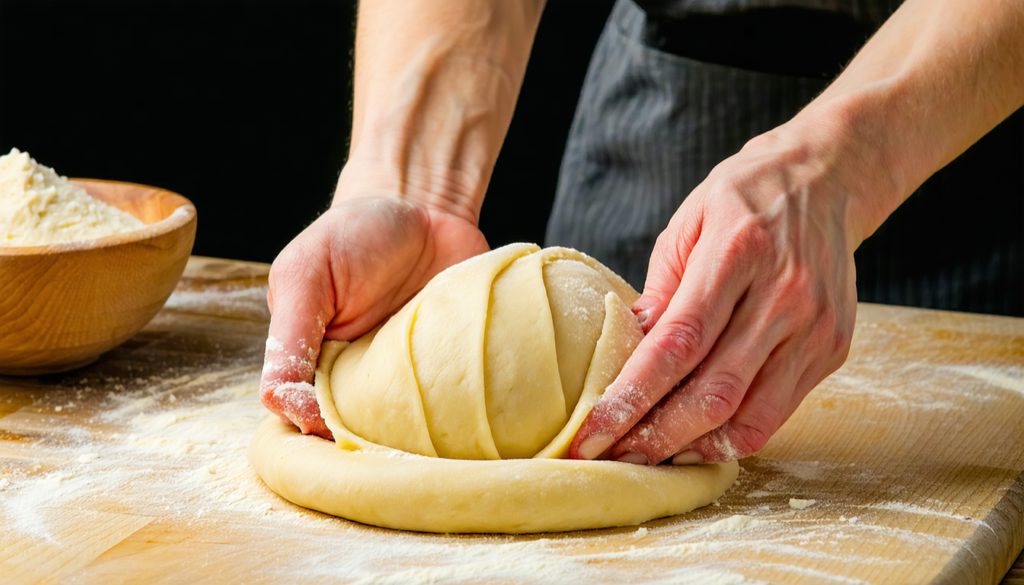}\\[1pt]
\includegraphics[width=1.2\linewidth]{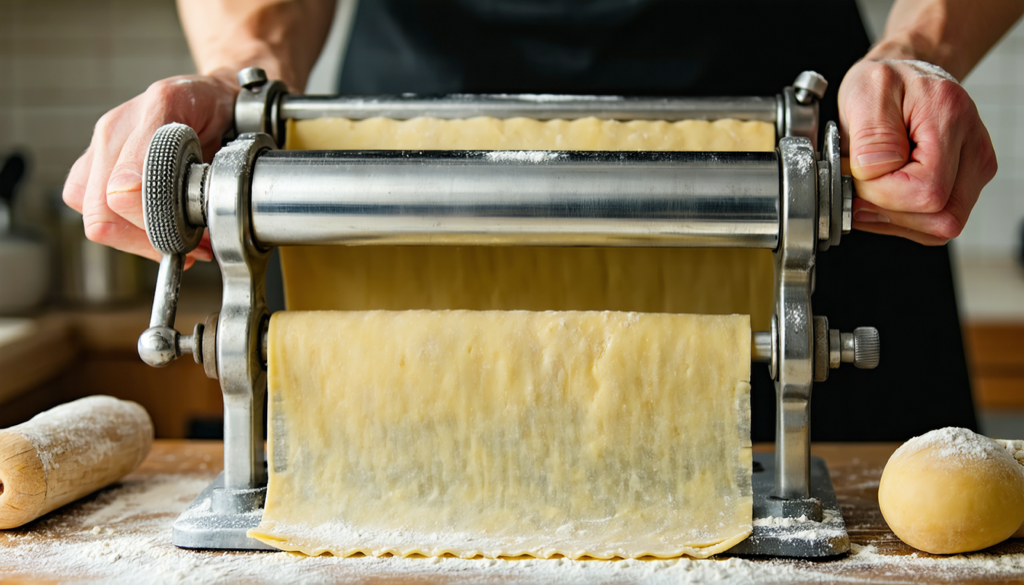}\\[1pt]
\includegraphics[width=1.2\linewidth]{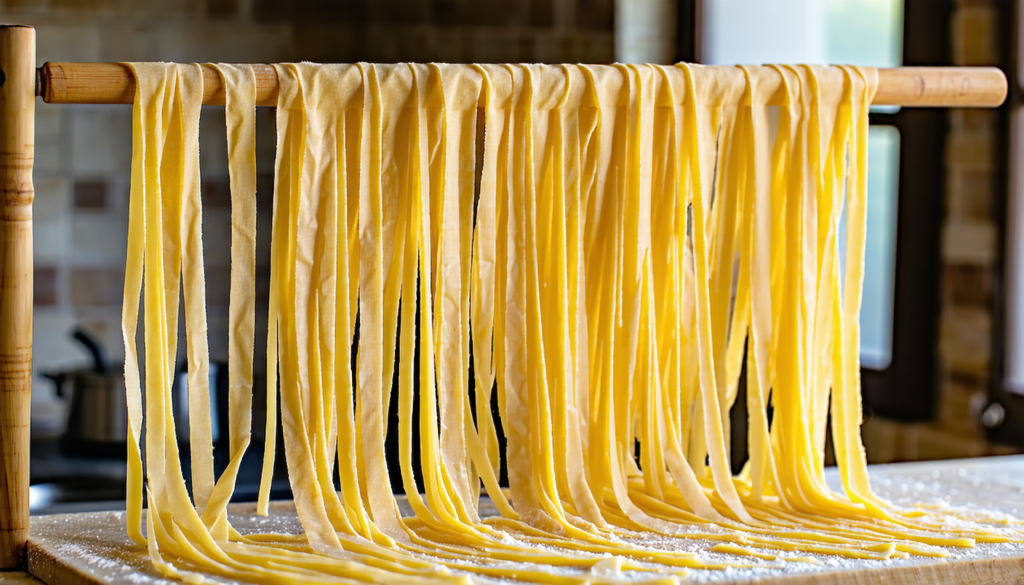}
\end{minipage}
\caption{Dataset construction pipeline. \textit{Left:} A podcast excerpt (555 words) with rich visual descriptions is extracted from SPoRC. \textit{Middle:} Claude Sonnet 4.5 generates five scene-specific image prompts. \textit{Right:} Stable Diffusion 3.5 renders the prompts into coherent visual sequences. This process creates 4,000 image-dialogue training pairs.}
\label{fig:pipeline}
\end{figure*}

Our dataset construction pipeline operates in three sequential stages, as illustrated in \cref{fig:pipeline}. First, we perform excerpt extraction by filtering the 1.1 million episodes in the SPoRC corpus to isolate high-quality, two-speaker (Host/Guest) dialogues. We used Claude Sonnet 4.5 to identify segments within these dialogues containing rich visual descriptions---scenes that listeners could vividly imagine. From over 81,000 suitable episodes, we extracted 4,000 high-quality excerpts averaging 900--1,100 words each.

Next, we proceed to image prompt generation. For each excerpt, Claude Sonnet 4.5 generated five detailed image prompts capturing the key visual scenes described in the dialogue. For example, a description of making pasta led to a prompt such as: \textit{``Hands covered in sticky, shaggy pasta dough on a wooden board, bits of white flour scattered around.''} These prompts were designed to be specific enough for high-fidelity image generation while remaining faithful to the podcast content. Finally, we execute image synthesis using Stable Diffusion 3.5 via AWS Bedrock. This process produced five images per excerpt, resulting in a dataset of 4,000 samples with approximately 20,000 total images.

Each training sample consists of a sequence of images paired with the original podcast transcript excerpt. While most samples contain five images, a small percentage (6.6\%) contain fewer due to Stable Diffusion's safety filters blocking prompts containing restricted concepts (e.g., violence, explicit content), so only 93.4\% of samples have exactly 5 images.

\cref{tab:image_dist} and \cref{tab:word_dist} show the distribution of our dataset. The majority of samples contain 5 images and 1,000--1,200 words, which provide substantial context for learning natural conversational patterns.

\begin{table}[t]
\centering
\begin{tabular}{@{}ccc@{}}
\toprule
\textbf{\# Images} & \textbf{Samples} & \textbf{\%} \\ \midrule
2 & 7 & 0.2 \\
3 & 31 & 0.8 \\
4 & 223 & 5.6 \\
5 & 3,737 & 93.4 \\
6 & 2 & 0.1 \\ \midrule
\textbf{Total} & \textbf{4,000} & \textbf{100} \\ \bottomrule
\end{tabular}
\caption{Image count distribution. 93.4\% of samples contain exactly 5 images.}
\label{tab:image_dist}
\end{table}

\begin{table}[t]
\centering
\begin{tabular}{@{}ccc@{}}
\toprule
\textbf{Word Range} & \textbf{Count} & \textbf{\%} \\ \midrule
0--500 & 131 & 3.3 \\
500--600 & 191 & 4.8 \\
600--700 & 251 & 6.3 \\
700--800 & 443 & 11.1 \\
800--900 & 537 & 13.4 \\
900--1,000 & 619 & 15.5 \\
1,000--1,200 & 1,104 & 27.6 \\
1,200--1,500 & 604 & 15.1 \\
1,500--2,000 & 111 & 2.8 \\
2,000+ & 9 & 0.2 \\ \midrule
\textbf{Total} & \textbf{4,000} & \textbf{100} \\ \bottomrule
\end{tabular}
\caption{Word count distribution. Peak at 1,000--1,200 words (27.6\%).}
\label{tab:word_dist}
\end{table}

The source SPoRC transcripts show key conversational characteristics. A speaker switch rate of 14.8 turns per 1,000 words, an average of 67.3 words per turn, and a mean transcript length of 975.8 words. These patterns, such as frequent speaker alternation and moderate turn lengths, define natural podcast dialogue and represent the conversational style our model learns to emulate.

This approach forces the model to learn not just \textit{what} is in the images, but \textit{how} two humans would discuss them dynamically, incorporating interruptions, reactions, and subjective opinions.

\section{Methodology}

\subsection{Model Architecture}
We selected Qwen3-VL-32B as our base model due to its balance of performance and efficiency. Unlike the 235B model, the 32B variant is fine-tunable on standard cluster hardware (e.g., A100 GPUs) while still retaining strong reasoning capabilities.

\subsection{Fine-Tuning Strategy}
We employed Low-Rank Adaptation (LoRA)~\cite{hu2022lora} to efficiently update the model weights while preserving pre-trained knowledge. Our training configuration is detailed in \cref{tab:hyperparams}. The vision encoder was frozen to prevent catastrophic forgetting of general visual features, while the LLM backbone (Attention and MLP modules) was fine-tuned with a rank $r=16$ and alpha $\alpha=32$. We used a conservative learning rate of $4e-6$ with cosine scheduling, weight decay of 0.1, and gradient clipping at 0.3 to stabilize training on multimodal inputs. Memory efficiency was managed using DeepSpeed ZeRO-3~\cite{rajbhandari2020zero}.

To enhance robustness and prevent repetition loops—a common failure mode in open-ended VLM generation—we applied NEFTune noise~\cite{jain2024neftune} ($\alpha=5.0$) to the embeddings and explicitly included the \texttt{<|im\_end|>} token in all training targets. This ensures the model learns clear termination signals for long-form narrative generation.

\begin{table}[t]
\centering
\begin{tabular}{@{}ll@{}}
\toprule
\textbf{Parameter} & \textbf{Value} \\ \midrule
\multicolumn{2}{@{}l}{\textit{LoRA Configuration}} \\
\quad LoRA Rank ($r$) & 16 \\
\quad LoRA Alpha ($\alpha$) & 32 \\
\quad LoRA Dropout & 0.05 \\ \midrule
\multicolumn{2}{@{}l}{\textit{Optimization}} \\
\quad Learning Rate & 4e-6 \\
\quad Batch Size (per GPU) & 1 \\
\quad Gradient Accumulation & 4 \\
\quad Effective Batch Size & 32 \\
\quad Epochs & 1 \\
\quad Weight Decay & 0.1 \\
\quad Warmup Ratio & 0.1 \\
\quad Max Gradient Norm & 0.3 \\
\quad LR Scheduler & Cosine \\ \midrule
\multicolumn{2}{@{}l}{\textit{Regularization \& Infrastructure}} \\
\quad NEFTune Noise Alpha & 5.0 \\
\quad Model Max Length & 8192 \\
\quad Precision & bf16 \\
\quad DeepSpeed & ZeRO-3 \\
\quad Gradient Checkpointing & True \\ \bottomrule
\end{tabular}
\caption{Fine-tuning configuration. We use conservative hyperparameters (low learning rate, high NEFTune noise) to prevent overfitting. Training completed in $\sim$12 hours on 8$\times$A100 GPUs.}
\label{tab:hyperparams}
\end{table}

\section{Experiments}

\subsection{Implementation Details}
Fine-tuning was conducted on a server with 8$\times$ NVIDIA A100 80GB GPUs using DeepSpeed ZeRO-3 for memory-efficient distributed training. The complete training run over 4,000 samples completed in approximately 11 hours (40,138 seconds), equivalent to 88 A100-GPU-hours, processing 126 optimization steps with an effective batch size of 32. Peak GPU memory utilization was approximately 75GB per device. A conservative learning rate of $4 \times 10^{-6}$ was chosen to ensure stable convergence on multimodal inputs, as higher rates caused gradient instability in early experiments. The final training loss converged to 2.56.

For text generation, we used nucleus sampling with a temperature of 0.7 and a top-p value of 0.8. The maximum number of new tokens was set to 2048 to allow for long-form podcast transcripts.

We evaluated our fine-tuned 32B model against the Qwen3-VL-235B (Base) model. Evaluation was conducted on a hold-out set of 50 real-world photo sequences from the VIST dataset~\cite{huang2016visual} (approx. 50,000 generated words) for both style metrics and visual grounding metrics. This strictly tests out-of-distribution generalization, as the model was trained exclusively on synthetic images.

\subsection{Metrics}
We assess performance across four dimensions. For visual grounding, we use CLIPScore~\cite{radford2021learning} ($w=2.5$) to measure the semantic similarity between the generated transcript and the input images. For conversational style, we measure \textit{average turn length} (words per speaker turn) and \textit{switch rate} (speaker changes per 1,000 words) to quantify the narrative flow. For lexical diversity, we use Distinct-n~\cite{li2016diversity} (specifically Distinct-2) to measure vocabulary richness, calculated as the ratio of unique n-grams to total n-grams. Finally, for naturalness, we employ an AI-as-a-Judge protocol: a pairwise comparison using three frontier models (Gemini 3 Pro, Claude Opus 4.5, and GPT 5.2) to determine which transcript sounds more like a ``natural, unscripted conversation.''

\subsection{Quantitative Results}

\cref{tab:results} presents the comparison between our fine-tuned model and the base model.

\begin{table}[h]
\centering
\small
\begin{tabular}{@{}lrrr@{}}
\toprule
\textbf{Metric} & \textbf{32B} & \textbf{235B} & \textbf{$\Delta$} \\ \midrule
\textit{Visual Grounding} & & & \\
CLIPScore~\cite{radford2021learning} & 20.39 & 20.39 & 0.00 \\ \midrule
\textit{Conversational Style} & & & \\
Avg. Turn Length & \textbf{57.5} & 38.0 & +51\% \\
Switch Rate (/1k words) & \textbf{16.0} & 27.0 & -41\% \\
Number of Turns & \textbf{15.8} & 24.5 & -35.5\% \\ \midrule
\textit{Diversity} & & & \\
Distinct-2 & 0.82 & \textbf{0.85} & -0.04 \\ \bottomrule
\end{tabular}
\caption{Comparison of narrative style and visual grounding. The 32B fine-tuned model generates longer, more substantial narrative turns while maintaining visual grounding.}
\label{tab:results}
\end{table}

\begin{table}[h]
\centering
\small
\begin{tabular}{@{}lcc@{}}
\toprule
\textbf{Metric} & \textbf{32B Finetuned} & \textbf{235B Base} \\ \midrule
Avg. Word Count & 972.1 $\pm$ 83.1 & 918.6 $\pm$ 51.7 \\
Avg. Gen Time (s) & 266.6 $\pm$ 31.6 & 475.0 $\pm$ 25.9 \\ \bottomrule
\end{tabular}
\caption{Generation statistics. The fine-tuned 32B model runs on 1$\times$A100 while the 235B model requires 8$\times$A100, yet the smaller model is 1.8$\times$ faster while producing slightly longer outputs.}
\label{tab:gen_stats}
\end{table}

Remarkably, our fine-tuned 32B model achieved an identical CLIPScore (20.39) to the 235B teacher model (\cref{tab:results}). This confirms that our efficient fine-tuning strategy successfully transferred narrative style without overfitting to the synthetic training domain, maintaining robust perception on real-world photographs. Additionally, \cref{tab:gen_stats} shows that our fine-tuned model running on a single A100 GPU is 1.8$\times$ faster than the 235B model on 8$\times$A100, while producing slightly longer outputs.

The most significant shift occurred in narrative structure. The 235B Base model tended toward short, ``ping-pong'' descriptions with an average of 24.5 turns per story (Avg Turn: 38 words), resembling a Q\&A session. Our 32B model produced significantly longer, more substantial turns (15.8 per story at 57.5 words each), allowing for the development of anecdotes, humor, and narrative arcs essential for a podcast format.

We observed a slight decrease in Distinct-2 scores (0.85 $\rightarrow$ 0.82). While normally a negative signal in text generation, qualitative analysis suggests this reflects the successful adoption of natural human speech patterns, which naturally include repetitive fillers (e.g., ``you know,'' ``I mean'') that robotic descriptions lack.

\subsection{Qualitative Evaluation: AI-as-a-Judge}

To assess ``Naturalness,'' which automated metrics miss, we employed a panel of three frontier models---Gemini 3 Pro, Claude Opus 4.5, and GPT 5.2---as expert judges. Each judge independently reviewed the 50 inference pairs using a rubric focused on conversational flow, reaction speed, and personality hallucination. Averaged across the 50 test samples and three judges, our model achieved a win rate of $>$80\%.

All three judges independently declared the 32B Fine-tuned model the decisive winner. The panel unanimously agreed that the 32B model produces natural ``podcast banter,'' while the 235B Base model functions more like a ``scripted narrator.'' \cref{tab:ai_judge_consensus} summarizes the key differences identified by the judges.

\begin{table*}[h]
\centering
\small
\setlength{\tabcolsep}{6pt}
\begin{tabular}{@{}p{0.12\linewidth}p{0.41\linewidth}p{0.41\linewidth}@{}}
\toprule
\textbf{Criterion} & \textbf{32B Fine-tuned (Ours)} & \textbf{235B Base (Baseline)} \\ \midrule
\textbf{Interruptions} & \textit{Natural Overlap.} Speakers naturally cut each other off (e.g., ``Wait, wait—'') and interrupt mid-thought, mimicking the chaotic rhythm of real speech. & \textit{Sequential Turns.} Speakers politely wait for complete paragraphs to end before responding, often resulting in a formal, less dynamic flow. \\ \addlinespace
\textbf{Reaction Speed} & \textit{Visceral \& Immediate.} Reactions (``No way!'', ``Hah!'') happen instantly driving the banter forward rather than reflecting on the previous point. & \textit{Delayed \& Considered.} Speakers describe the visual evidence first, then offer a measured, polite response, slowing down the conversational flow. \\ \addlinespace
\textbf{Phrasing} & \textit{Casual \& Imperfect.} Uses fillers (``kinda'', ``sorta''), sentence trails, and uneven pacing to mimic spontaneous thought processes. & \textit{Polished Prose.} Uses grammatically perfect, formal sentences that sound like they are being read from a prepared script or essay. \\ \addlinespace
\textbf{Personality} & \textit{Creative Hallucination.} Invents vivid anecdotes (e.g., the ``sandwich smuggler'') and character backstories to explain visual details. & \textit{Observational Safety.} Sticks strictly to describing visible elements. Analysis is visually accurate but lacks character voice or narrative invention. \\ \midrule
\textbf{Verdict} & \textit{Unanimous Winner (3/3 Judges)} & \textit{Preferred only for Captioning/Audiobooks} \\ \bottomrule
\end{tabular}
\caption{Consensus evaluation from the AI-as-a-Judge panel. All three judges (Gemini 3 Pro, Claude Opus 4.5, GPT 5.2) favored the 32B model for its ability to break the rigid structure of standard LLM generation and hallucinate plausible host personas.}
\label{tab:ai_judge_consensus}
\end{table*}

The judges consistently highlighted a key structural difference: the 235B Base model adheres to a rigid ``turn-taking'' protocol, resulting in what Claude Opus 4.5 described as a ``scripted panel.'' In contrast, the 32B Fine-tuned model learned to break this pattern by introducing interruptions and non-sequiturs that mimic human spontaneity. GPT 5.2 noted that our model ``sounds like people with microphones thinking out loud,'' and prioritizes entertainment value over the literal visual accuracy favored by the base model.

\subsection{Qualitative Examples}

The input prompts and full generated transcripts for the 32B Base, 32B Fine-tuned, and 235B Base models are provided in \cref{sec:appendix}.

\cref{fig:qualitative} presents representative outputs from the three models on the same wedding photo sequence. The 32B Base model (see \cref{sec:appendix}) serves as a baseline. Although it generates conversational text, it lacks specific ``host'' personas and often reverts to generic image description (e.g., ``I see the third photo...''). The 235B Base model produces high-quality text but remains tethered to the visual input. Its generation amounts to a sophisticated image caption ("the wood paneling, the chandeliers"). In contrast, our 32B Fine-tuned model successfully "hallucinates personality," inventing a backstory about a "sandwich guy" and a drunk best man quoting Shakespeare. This creative leap that moves from visual evidence to plausible social context is the hallmark of engaging storytelling. The fine-tuned model also adopts a "meta-commentary" style ("So I was thinking, what if we did a podcast..."). It frames the narrative as a planned episode rather than a reactive description. This structural shift, combined with the use of specific humor ("Cupcake Catastrophes"), demonstrates that the model has learned the genre of podcasting, not just the task of description.

\begin{figure*}[t]
\centering
\scriptsize
\setlength{\tabcolsep}{3pt}
\begin{tabular}{@{}p{0.31\textwidth}|p{0.31\textwidth}|p{0.31\textwidth}@{}}
\textbf{32B Base} & \textbf{32B Fine-tuned (Ours)} & \textbf{235B Base} \\
\hline\\[-6pt]
\textit{Speaker 1: Hey, Jamie! You see these photos I sent over? I was going through some old wedding pics from last year, and I just got hit with a wave of nostalgia. I mean, look at this first one---five guys in suits, standing outside some grand old building, laughing. It's like they're plotting something. Or maybe just sharing a joke before the ceremony.}

\textit{Speaker 2: Oh, I love that one! There's something so authentic about it. You can tell they're not posing---they're just there, in the moment. And that guy with the camera bag? He's probably the photographer, right? But he's laughing too, like he's part of the crew, not just the observer. I think that's what I love about wedding photos---they capture the real, unscripted stuff.}

\textit{Speaker 1: Exactly! And look at the building---it's got that classic stone facade, the little decorative windows. Feels like a courthouse or maybe a town hall. I remember the ceremony was in a council chamber---that second photo. You can see the wood paneling, the green benches, the little microphones on each desk. It's so formal, but then you've got people in hats and dresses, and someone's taking a photo with their phone. It's like, ``We're in a courtroom, but we're celebrating love!'}

\textit{......}
&
\textit{Speaker 1: So I was thinking, what if we did a podcast about weddings? Not like, the whole ``how to plan your dream day'' kind of thing, but more like, the real stories behind them. The weird stuff, the funny moments, the things you'd never tell your grandma.}

\textit{Speaker 2: Oh, I'm in. I've been to enough weddings to know that 90\% of them are just people pretending to be happy while someone else's life is being celebrated. But the other 10\%? That's gold. Like that time I saw a guy in a full suit trying to sneak a sandwich into the ceremony. He was like, ``I'm not hungry, I'm just... carrying it for my mom.''}

\textit{Speaker 3: Hah! That's classic. I remember one wedding where the best man got so drunk during the speeches he started quoting Shakespeare, but he kept mixing up Hamlet and Macbeth. He was like, ``To be or not to be... wait, no, that's the one with the witches, right?'' The whole room was silent. Then he looked at the bride and said, ``You're my Juliet, and I'm your... uh... Romeo? Or was it Mercutio? I'm confused.''}

\textit{......}
&
\textit{Speaker 1: Alright, so I'm still trying to process how we ended up at a wedding in a courtroom. Like, seriously? Was that even legal? I mean, the wood paneling, the chandeliers, the judge's bench---it felt like we were about to get sentenced, not celebrate.}

\textit{Speaker 2: It was totally legal! The registrar was legit. And honestly? It was kind of brilliant. No boring church, no awkward hymns---just a really elegant room with a weirdly regal vibe. Plus, the lighting? Chef's kiss. You could see every detail---the flowers, the hats, even the guy in the front row trying to sneak a selfie with his phone.}

\textit{Speaker 3: That guy was me. And I got a solid shot of the bride walking in. She looked amazing. That veil, the dress---it was like she stepped out of a vintage magazine. And the groom? He was beaming. You could tell he was trying not to cry, but his eyes were definitely glistening.}

\textit{Speaker 4: I think everyone was trying not to cry. Even the guy with the mustache who was sitting next to me---he was wiping his eyes with his napkin. And then he started recording the whole thing on his phone like he was making a documentary. I swear, I thought he was going to stand up and yell ``Objection!'' halfway through.}

\textit{......}
\\
\end{tabular}
\caption{Qualitative comparison of generated transcripts on the same wedding photo sequence (\cref{fig:teaser}). \textit{32B Base:} Conversational but observational, directly describing photos (``look at this first one,'' ``that second photo''). \textit{32B Fine-tuned:} Most natural style---speakers riff on meta-topics, share personal anecdotes (the sandwich guy, the drunk best man), and exhibit genuine humor without referencing images. \textit{235B Base:} Engaging but overly descriptive---phrases like ``the wood paneling, the chandeliers, the judge's bench'' read more like captions than natural speech.}
\label{fig:qualitative}
\end{figure*}

\subsection{Generalization to Diverse Domains}
To ensure our model's stylistic gains were not limited to sentimental domains like weddings, we analyzed its performance on high-energy, dynamic narratives. In \cref{fig:teaser} (Row 2), the visual input consists of action shots involving motorcycles and marathon running.

The 235B Base model tends to treat the images as separate events, jumping abruptly from one topic to the next: \textit{``Okay, but let's talk about the marathon.''} It accurately identifies the activities but fails to weave them into a cohesive character arc.

In contrast, the 32B Fine-tuned model connects the disparate visual elements through a consistent character narrative. It transitions from the motorcycle discussion to the marathon by analyzing the subject's personality: \textit{``You've always been the kind of person who just goes for it... I remember when you signed up for that marathon.''} This demonstrates that the fine-tuned model acts as a \textit{narrative filter}, prioritizing character consistency over simple visual enumeration.

Beyond domain generalization, a distinct, high-level behavior emerging from our fine-tuning is the hallucination of shared history. While standard VLMs typically ground their dialogue strictly in the \textit{present} visual evidence (e.g., \textit{``Look at that guy''}), our 32B model frequently frames the narrative as a \textit{recollection} of a past shared between the speakers. For instance, when processing \cref{fig:teaser} (Row 2), the 32B model generates the line: \textit{``I remember you texting me during the race, saying `I can't believe I'm doing this!'''}. This specific detail, like a text message sent during a past event, cannot be inferred from the pixels of the image. By hallucinating this shared memory, the model creates a layer of synthetic intimacy. This successfully mimics the rapport found in real podcasts, where hosts often reference off-mic interactions or past experiences that the audience is not privy to, thereby deepening the listener's immersion.

\section{Limitations and Societal Impact}

Despite our fine-tuned model's improved narrative depth, it relies on the quality of the underlying SPoRC transcripts. We observed that the model occasionally hallucinates visual details not present in the image if they strongly fit the ``podcast persona'' (e.g., describing a smell or sound). Furthermore, our method currently processes a fixed sequence of five images; extending this to variable-length or streaming inputs remains future work. While our AI-as-a-Judge framework provides a scalable solution for qualitative assessment, the evaluation of ``naturalness'' and ``narrative quality'' remains inherently subjective.

\vspace{0.5em}
While visual podcast generation offers creative potential, it presents risks regarding impersonation and information integrity. Our model is designed for entertainment, but the capability to synthesize believable dialogue from images could be misused to fabricate events or misrepresent individuals. Additionally, generated narratives may inherit biases present in the podcast training data, and potentially reinforce stereotypes related to gender, accent, or dialect. Future deployments should incorporate watermarking to distinguish between authentic human testimony and AI-generated content.

\section{Conclusion}

In this work, we demonstrated that narrative quality in Vision-Language Models can be significantly improved through targeted fine-tuning on high-quality spoken dialogue data. Our empirical results demonstrate the efficacy of this method. Despite being nearly 8x smaller than the base model, our fine-tuned 32B variant achieved a $>$80\% win rate in AI-judge pairwise evaluations for conversational naturalness. Quantitative metrics further confirm this stylistic shift: the model increased the average turn length by 51\% (from 38.0 to 57.5 words) and reduced the speaker switch rate by 41\% (from 27.0 to 16.0 per 1k words), successfully replicating the substantive, multi-sentence flow of authentic podcasts. Importantly, this narrative enrichment was achieved without hallucinating incorrect visual facts, as the model maintained an identical CLIPScore of 20.39 compared to the teacher model, while also offering a 1.8$\times$ speedup in inference latency. We proved that models trained on high-quality synthetic image-text pairs can effectively generalize to real-world visual storytelling, suggesting that style metrics such as turn length and speaker switch rate provide a more valuable signal than traditional n-gram overlaps for creative tasks.

Future work will focus on two critical avenues. First, we aim to extend the input to support variable-length image sequences and real-time streaming image frames or videos. Second, we plan to enhance the text-to-audio interface by training the model to output structured prosody markers, such as Speech Synthesis Markup Language (SSML) tags or explicit emotion tokens (e.g., \texttt{(excited)}, \texttt{[laughter]}). This would allow the VLM to explicitly direct the intended emotional nuance, pacing, and tone shifts of the generated script.

{\small
\bibliographystyle{unsrtnat}
\bibliography{main}
}

\appendix

\section{Inference Prompt}
\label{sec:inference_prompt}

For all inference experiments, we provide the model with the following system prompt along with the five input images:

\begin{quote}
\textit{``Generate a natural conversational podcast dialogue. Use the format Speaker 1:, Speaker 2:, Speaker 3:, etc. for multiple speakers. Do not reference the images or use phrases like `our first image'. Write casual, authentic spoken dialogue without introductions or sign-offs. The word count should be around 800 words.''}
\end{quote}

This prompt was designed to elicit natural podcast-style dialogue while preventing the model from producing meta-commentary about the images themselves.

\section{Full Transcript Comparison}
\label{sec:appendix}

Here we present the full, unedited transcripts for both evaluation examples shown in Figure 1 of the main paper, as generated by the 32B Base, 32B Fine-tuned (Ours), and 235B Base models. Sentences that directly reference visual content from the input images are highlighted in {\color{red} red text}, with the corresponding image index noted in parentheses.

\subsection*{Example 1: Wedding Party (Story ID: 50242)}
This corresponds to Figure 1, Row 1 in the main paper.

\subsection{Qwen3-VL 32B Base Model (689 words)}

\textbf{Alex:} Hey, Jamie! You see these photos I sent over? I was going through some old wedding pics from last year, and I just got hit with a wave of nostalgia. {\color{red} I mean, look at this first one---five guys in suits, standing outside some grand old building, laughing. (Image 1)} It's like they're plotting something. Or maybe just sharing a joke before the ceremony.

\textbf{Jamie:} Oh, I love that one! There's something so authentic about it. You can tell they're not posing---they're just there, in the moment. {\color{red} And that guy with the camera bag? (Image 1)} He's probably the photographer, right? But he's laughing too, like he's part of the crew, not just the observer. I think that's what I love about wedding photos---they capture the real, unscripted stuff.

\textbf{Alex:} Exactly. {\color{red} And then you see the third photo---the bride and groom outside, surrounded by friends. She's got that little net veil, and he's in a grey vest with his sleeves rolled up. (Image 3)} It's so relaxed, so modern, but still elegant. {\color{red} And they're all holding those blue plastic cups---I mean, come on, that's such a fun detail. (Image 3)} It's not fancy champagne flutes; it's real celebration.

\textbf{Jamie:} Oh, I love that! It's like, ``We're not pretending to be fancy. We're just happy.'' {\color{red} And the greenery behind them---it's lush, natural. It feels like they're in a garden, maybe a courtyard. (Image 3)} It's intimate, not staged.

\textbf{Alex:} {\color{red} And then you get to the cupcakes---the fifth photo. Oh my god, look at those! They're on these elegant tiered stands, but they're cupcakes. (Image 5)} Not a big cake. {\color{red} And they're all different---chocolate, vanilla, pink frosting, swirls. (Image 5)} It's like a dessert buffet.

\textbf{Jamie:} I can! It's like the perfect end to the day---sweet, colorful, and shared. And the fact that they chose cupcakes instead of a traditional cake? That's such a modern touch.

\textbf{Alex:} Exactly. And I think that's the theme that ties all these photos together---authenticity. {\color{red} Whether it's the guys laughing outside the building (Image 1)}, the bride in her simple dress, {\color{red} the chaotic reception (Image 4)}, or {\color{red} the cupcakes (Image 5)}---it's all about being true to who you are.

\textbf{Jamie:} Well said. Here's to messy, beautiful, real moments---and to cupcakes.

\textbf{Alex:} Cheers to that.

\subsection{Qwen3-VL 32B Fine-tuned (Ours) (1,191 words)}

\textbf{Speaker 1:} So I was thinking, what if we did a podcast about weddings? Not like, the whole ``how to plan your dream day'' kind of thing, but more like, the real stories behind them. The weird stuff, the funny moments, the things you'd never tell your grandma.

\textbf{Speaker 2:} Oh, I'm in. I've been to enough weddings to know that 90\% of them are just people pretending to be happy while someone else's life is being celebrated. But the other 10\%? That's gold. Like that time I saw a guy in a full suit trying to sneak a sandwich into the ceremony. He was like, ``I'm not hungry, I'm just... carrying it for my mom.''

\textbf{Speaker 3:} Hah! That's classic. I remember one wedding where the best man got so drunk during the speeches he started quoting Shakespeare, but he kept mixing up Hamlet and Macbeth. He was like, ``To be or not to be... wait, no, that's the one with the witches, right?'' The whole room was silent. Then he looked at the bride and said, ``You're my Juliet, and I'm your... uh... Romeo? Or was it Mercutio? I'm confused.''

\textbf{Speaker 1:} I've got one that's even better. {\color{red} My cousin got married at this super fancy old courthouse. (Image 2)} Everyone was dressed up, the whole thing was formal. But then, halfway through the ceremony, {\color{red} this guy walks in with a giant camera and a backpack, starts snapping photos like he's covering a war zone. (Image 1)} And he's not even part of the wedding party.

\textbf{Speaker 2:} Wait, wait, wait---did he get kicked out?

\textbf{Speaker 1:} No! The bride's brother actually invited him. Turns out he's a freelance photographer and he'd been following the couple for months, documenting their ``love story'' without them knowing. The bride was like, ``Wait, you've been stalking us?'' And he was like, ``Stalking? I was \textit{documenting}!''

\textbf{Speaker 3:} That's insane. But also kind of sweet, in a creepy way.

\textbf{Speaker 2:} Speaking of weird, I once went to a wedding where the reception was in a barn, but {\color{red} they had a full-on cupcake tower. Like, three tiers, 50 cupcakes, all different flavors. (Image 5)} And someone had clearly forgotten to tell the caterer that the bride was allergic to nuts. So when they cut into the first one, it was like, ``Oh no, it's got almonds in it!'' And the bride's face turned white. She didn't say anything, just handed it to the groom and said, ``You eat it. I'm not risking anaphylaxis for a wedding cake.''

\textbf{Speaker 1:} That's terrifying. But also, I love that she didn't panic. She just handed it off like, ``You're the one who said `I do'---you deal with it.''

\textbf{Speaker 3:} I've got a story that's more about the aftermath. My friend got married in this tiny village hall, and after the ceremony, they had a little reception with sandwiches and wine. But the thing was, {\color{red} the groom's dad had brought his laptop. Not for work. For the \textit{reception}. He was sitting at a table, typing away, completely ignoring everyone. (Image 4)} We were like, ``Dad, are you working?'' And he was like, ``No, I'm just... updating my LinkedIn. You never know when you'll need it.''

\textbf{Speaker 2:} That's the most relatable thing I've ever heard.

\textbf{Speaker 1:} {\color{red} And then there's the one where the bride wore a veil, but it was like a tiny birdcage thing. And she was standing next to the groom, who was in a vest, (Image 3)} and they were both looking at the guests like, ``Is this happening?'' And then she took a sip of water and the veil got caught on the rim of the glass.

\textbf{Speaker 3:} I've seen that. I've seen people trip on their train, spill wine on the groom, forget their vows... it's chaos. But the best part is, no one really cares. Everyone's just happy to be there. That's what makes weddings so beautiful. It's not perfect. It's human.

\textbf{Speaker 2:} Exactly. I think that's why I love them. It's not about the dress or the venue or the cake. It's about the people.

\textbf{Speaker 1:} So we're doing it? The ``Weird Weddings'' podcast?

\textbf{Speaker 2:} I'm in. Let's call it ``The Unplanned Vows.''

\textbf{Speaker 3:} I like it. And we can have a segment called ``Cupcake Catastrophes.''

\textbf{Speaker 1:} Perfect. We'll start with the courthouse photographer story. That's a solid opener.

\subsection{Qwen3-VL 235B Base Model (979 words)}

\textbf{Speaker 1:} Alright, so I'm still trying to process how we ended up at a wedding in a courtroom. Like, seriously? Was that even legal? {\color{red} I mean, the wood paneling, the chandeliers, the judge's bench with the green curtain behind it (Image 2)}---it felt like we were about to get sentenced, not celebrate.

\textbf{Speaker 2:} It was totally legal! The registrar was legit. And honestly? It was kind of brilliant. No boring church, no awkward hymns---just a really elegant room with a weirdly regal vibe. Plus, the lighting? Chef's kiss. You could see every detail---the flowers, the hats, {\color{red} even the guy in the front row trying to sneak a selfie with his phone. (Image 2)}

\textbf{Speaker 3:} That guy was me. And I got a solid shot of the bride walking in. She looked amazing. {\color{red} That veil, the dress---it was like she stepped out of a vintage magazine. (Image 3)} And the groom? He was beaming. You could tell he was trying not to cry, but his eyes were definitely glistening.

\textbf{Speaker 4:} I think everyone was trying not to cry. {\color{red} Even the guy with the mustache who was sitting next to me---he was wiping his eyes with his napkin. And then he started recording the whole thing on his phone like he was making a documentary. (Image 2)} I swear, I thought he was going to stand up and yell ``Objection!'' halfway through.

\textbf{Speaker 1:} Honestly, I think that would've been the highlight. ``Objection! The groom is too handsome!'' But seriously, the whole thing was so... intentional. Like, they didn't just pick a venue---they curated an experience. Even the reception was a vibe. {\color{red} That checkered tablecloth? The Union Jack bunting? (Image 4)} It was like a British garden party meets a hipster wedding.

\textbf{Speaker 2:} And don't even get me started on the food. I mean, I was expecting a few sandwiches and some cheap wine, but no---{\color{red} they had this whole spread. Cheese boards, little sandwiches, and then the cupcakes. (Image 4, 5)} {\color{red} Oh my god, the cupcakes. They were stacked on these tiered stands like they were in a fancy bakery window. (Image 5)} {\color{red} Chocolate, vanilla, pink swirls (Image 5)}---I think I had three.

\textbf{Speaker 3:} You had five. I counted. And you were trying to be sneaky about it, but I saw you sneak back for that chocolate one with the pink frosting. You're lucky the bride didn't catch you---she was busy trying to keep her veil from getting caught in her chair.

\textbf{Speaker 4:} That veil was a menace. I saw her trying to adjust it while she was talking to the guests. And then there was that moment when she was standing by the window, and the light hit her just right---it was like a movie scene. I actually stopped eating for a second to take it in.

\textbf{Speaker 1:} Same. I think we all did. It was one of those moments where you just pause and think, ``This is why we do this.'' You know? Weddings are weird---they're this mix of chaos and beauty, and somehow it all comes together.

\textbf{Speaker 2:} And let's not forget the groom's speech. I mean, he started off with a joke about how he used to think marriage was just a legal contract, and then he got all serious and said, ``But now I know it's about choosing to be happy with the same person every day.'' I swear, I choked on my wine.

\textbf{Speaker 3:} I think half the room choked on their wine. And then the best man had to step in and say, ``Alright, enough of that---let's get back to the cake!''

\textbf{Speaker 4:} I was waiting for someone to accidentally stab themselves with it. But nope---everyone was surprisingly coordinated. Even the little girl in the pink dress who was sitting next to me---she was eating her cake like a pro. No mess, no spills, just pure focus.

\textbf{Speaker 1:} That kid was a legend. I think she's going to be a wedding planner one day.

\textbf{Speaker 2:} You did spill your drink on your suit. Twice. But hey, at least you looked good doing it. And honestly, that's what weddings are about---looking good while you're making a mess. It's like a rite of passage.

\textbf{Speaker 3:} I think the real rite of passage was surviving {\color{red} the group photo. (Image 1)} I swear, we must've taken 20 pictures before everyone was smiling at the same time.

\textbf{Speaker 4:} But we pulled it off. And now we have these amazing photos to prove it. I mean, I've already sent them to my mom, and she's already planning her own wedding based on this one. She wants the courtroom, {\color{red} the cupcakes (Image 5)}, the checkered tablecloths---the whole shebang.

\textbf{Speaker 1:} Tell her to add {\color{red} the mustache guy with the phone. (Image 2)} He's a must-have.

\textbf{Speaker 2:} Agreed. And hey, if we're lucky, maybe we'll get invited to the next one. I hear they're thinking about doing it on a boat. With fireworks. And more cupcakes.

\textbf{Speaker 3:} I'm already packing my bag.

\subsection*{Example 2: Motorcycle \& Marathon (Story ID: 47306)}
This corresponds to Figure 1, Row 2 in the main paper.

\subsection{Qwen3-VL 32B Base Model (689 words)}

\textbf{Alex:} Hey, Jamie! You see those photos I sent over? {\color{red} The ones with the woman in the leather gear (Image 1)} and then the marathon stuff? I was flipping through them and I just had to talk about them. There's something really cool about how they tell a story---not just of one person, but of different sides of life, you know?

\textbf{Jamie:} Oh, I saw them! I was actually kind of struck by the contrast. Like, {\color{red} first you've got her in this full-on black leather jacket and pants, standing in a room, holding a helmet (Image 1)}, smiling like she's ready to take on the world. And then---boom---{\color{red} she's on a motorcycle, helmet on, looking like she's about to ride off into the sunset. (Image 2)} {\color{red} And the bike---it's a Yamaha, right? Not some cruiser. It's sporty, sleek. (Image 2)} She's not just riding for fun; she's riding with purpose.

\textbf{Alex:} And then---\textit{cut}---{\color{red} she's at a marathon. And she's not even in the same outfit! She's in this soft white turtleneck, which is so different from the leather. (Image 3)} It's almost like she's shedding that tough exterior for something more... approachable? Or maybe just more comfortable for running. But she's still smiling. Still confident. Just in a different way.

\textbf{Jamie:} Right? And the setting changes too. From a quiet room, to {\color{red} a parking lot with a car and a hedge (Image 2)}, to {\color{red} this huge, crowded grassy field with a giant ``Íslandsbanka Reykjavík Maraton'' banner. (Image 3)} It's like she's moving from private to public, from solitary to community.

\textbf{Jamie:} It's like the marathon becomes this big, messy, beautiful human moment. And she's part of it---not just as a runner, but as someone who's \textit{feeling} it. You can see it in her face in the first marathon photo---she's not just posing. She's \textit{there}. She's present. And then, {\color{red} in the running shot, she's laughing. (Image 4, 5)} That's the real win, isn't it? Not finishing first, but finishing with joy.

\textbf{Alex:} Absolutely. And I think that's the thread connecting all the photos---this sense of \textit{living fully}. Whether she's {\color{red} riding a motorcycle (Image 2)}, {\color{red} standing in her leather gear (Image 1)}, or {\color{red} running a marathon (Image 3, 4, 5)}, she's engaged. She's not just going through the motions. She's choosing to be in the moment, to embrace the challenge, to wear the gear that fits the moment.

\textbf{Jamie:} And maybe, just maybe, to laugh while you're doing it.

\textbf{Alex:} Always. Always laugh.

\subsection{Qwen3-VL 32B Fine-tuned (Ours) (935 words)}

\textbf{Speaker 1:} So I was looking through some old photos the other day, and I came across these pictures from a few years back. I was trying to remember what the heck I was doing, and then it all came back. I had just gotten my motorcycle license, and I was so excited. {\color{red} I remember buying this black leather jacket and matching pants. (Image 1)} I thought I looked like a total badass.

\textbf{Speaker 2:} That's awesome! I remember you showing me the jacket. You were so proud of it. {\color{red} You even had the whole outfit---the boots, the gloves, the helmet. (Image 1)} You looked like you were ready to take on the open road.

\textbf{Speaker 1:} I was! {\color{red} I had this little Yamaha (Image 2)}, and I'd ride around town just to feel the wind in my face. It was such a rush. I even did a little stunt once---just a tiny wheelie---and I almost wiped out. I laughed so hard afterward. I was more scared than I let on, but I felt so alive.

\textbf{Speaker 2:} You always did love that adrenaline, didn't you? I remember you telling me you wanted to ride across Europe someday. You even had a map on your wall with all the routes you wanted to take.

\textbf{Speaker 1:} I did! I still have that map. I didn't end up doing it, but I've been thinking about it lately. Maybe next year? I mean, I'm not getting any younger, and I don't want to wait too long.

\textbf{Speaker 3:} You know, I think you should. Life's too short to wait for the ``perfect'' time. You've always been the kind of person who just goes for it. I remember when you signed up for that marathon---out of nowhere---and you didn't even train for it. You just showed up and ran it.

\textbf{Speaker 1:} Ha! Yeah, that was a disaster. I think I ran the whole thing at a jog, and I was so sore for a week. But you know what? I loved it. I remember the day of the race---it was raining, and {\color{red} the streets were wet (Image 5)}, but there was this energy in the air. Everyone was so excited. {\color{red} I was wearing this white jacket (Image 3)}, and I felt like I was part of something big.

\textbf{Speaker 2:} You were! I remember you texting me during the race, saying, ``I can't believe I'm doing this!'' I was so proud of you. You didn't even train, and you finished. That's impressive.

\textbf{Speaker 3:} And you didn't just finish---{\color{red} you ran with this huge smile on your face the whole time. I saw you in the photos, and you looked like you were having the time of your life. (Image 3, 5)} I think that's what I love most about you---you find joy in the little things, even when they're challenging.

\textbf{Speaker 1:} Well, I guess I'm just a sucker for a good challenge. I mean, I didn't even know I liked running until I signed up for that race. I thought I'd hate it. But something about being out there, surrounded by people who were all pushing themselves---it just felt right.

\textbf{Speaker 2:} It's funny how sometimes we do things on a whim, and they end up changing us. I think that's what happened with you and the motorcycle, too. You weren't really into it before, but then you got your license, and suddenly it was your thing.

\textbf{Speaker 1:} Exactly. I think we all have those moments---that one thing that just clicks. For me, it was the freedom of the road and the rhythm of running. Both of them made me feel like I was truly living.

\textbf{Speaker 3:} You know, I think that's why you're such an inspiration to people. You don't wait for permission to do the things you want. You just go for it. And that's something a lot of us could learn from.

\textbf{Speaker 1:} Well, I didn't plan any of it. I just followed my curiosity. I remember thinking, ``What if I try this?'' and then I did. And most of the time, it turned out okay---sometimes even better than I expected.

\textbf{Speaker 2:} Like when you rode that motorcycle through the rain and got soaked, but you still grinned the whole way home.

\textbf{Speaker 1:} (laughs) I did! I was covered in mud, and my jacket was dripping, but I didn't care. I felt like I'd conquered the world. I think that's what it's all about---not being afraid to get messy, to take risks, to just... be.

\textbf{Speaker 3:} I think that's the real takeaway here. Life isn't about being perfect or having everything figured out. It's about showing up, even when you're scared, and just doing it. And you've done that more than most.

\textbf{Speaker 1:} Thanks, I guess. I just try to live in the moment. And hey, if I can inspire even one person to try something new---whether it's riding a motorcycle or running a marathon---then I've done something right.

\textbf{Speaker 2:} You've done more than that. You've reminded us all that we're capable of more than we think. And that's a pretty big deal.

\textbf{Speaker 3:} Absolutely. So here's to you---to the adventures, the mishaps, the rain-soaked runs, and the muddy rides. May you keep going, keep exploring, and keep smiling through it all.

\textbf{Speaker 1:} (smiling) I'll try my best. And who knows? Maybe next year, I'll be riding across Europe with that map finally coming to life.

\subsection{Qwen3-VL 235B Base Model (910 words)}

\textbf{Speaker 1:} You know what's wild? I used to think I'd never be the kind of person who'd {\color{red} wear head-to-toe leather (Image 1)} just to look cool. But then I got into motorcycling, and suddenly, it's not about looking cool---it's about not getting scraped off the pavement. {\color{red} That jacket? Saved my ass more than once. (Image 1)}

\textbf{Speaker 2:} {\color{red} Wait, you're the one who rides that silver Yamaha? (Image 2)} I remember seeing you {\color{red} parked next to that old blue truck. (Image 2)} You looked like you were about to take off for Reykjavík or something. Did you actually go anywhere epic?

\textbf{Speaker 1:} Ha! No, just to the grocery store. But I swear, riding through the city in full gear, people stare like you're some kind of outlaw. Honestly, I just want to get my oat milk without getting rained on. The helmet's the real MVP---it keeps the wind out of my ears and the bugs out of my mouth.

\textbf{Speaker 3:} Okay, but let's talk about the marathon. {\color{red} I saw you at the start line, all chill in that white fleece (Image 3)}, sunglasses on, like you were just out for a Sunday stroll. And then---bam---you're sprinting past people like you've got a personal vendetta against the finish line.

\textbf{Speaker 1:} {\color{red} That was the Reykjavík Marathon, right? (Image 3)} Yeah, I was feeling good that day. The crowd was insane---everyone cheering, waving, some guy even handed me a beer as I ran by. I didn't take it, obviously, but the gesture? Priceless. You should've seen the look on the guy's face when I waved and kept going.

\textbf{Speaker 2:} I was there too! {\color{red} I was one of the idiots running in the blue shirt with the red cap. (Image 4)} You passed me right after the hill near the park. I was wheezing like a broken accordion. How do you even train for that? Do you just... run everywhere?

\textbf{Speaker 1:} Pretty much. I run to the store, to work, to the bus stop---I even ran to my own birthday party once. My friends thought I was nuts. But honestly, once you get into it, it's addictive. The rhythm, the sweat, the way your legs just... keep going. Plus, it's a great way to see the city. You notice stuff you'd never see in a car.

\textbf{Speaker 3:} I tried running once. Made it two blocks before I had to stop and pretend I was looking at a map. I think my lungs are still mad at me. But hey, at least I can appreciate the view from the sidelines. Like, {\color{red} watching you guys run past those green buildings? That was a vibe. The crowd was so hyped---people were dancing, clapping, some even had signs. (Image 5)} One guy had a sign that said ``You're doing great, even if you're last!'' I cried a little.

\textbf{Speaker 1:} That's the spirit! Marathons aren't just about speed---they're about showing up, putting one foot in front of the other, and not letting the rain or the hills or the guy with the beer stop you. I mean, I've had days where I wanted to quit halfway, but then I'd see someone cheering in a ridiculous costume, and I'd be like, ``Okay, I can do this for five more minutes.''

\textbf{Speaker 2:} I love that. It's like, life's a marathon, right? You don't have to be the fastest, just the most stubborn. Speaking of stubborn---how's the motorcycle thing going? Still riding in all weather?

\textbf{Speaker 1:} Absolutely. Rain, snow, sleet---I've done it all. There's something about the wind hitting your face, the engine rumbling under you, that just clears your head. Plus, it's a great way to escape the city when you need a break. I'll just hop on, ride out to the coast, and watch the waves crash. It's my therapy.

\textbf{Speaker 3:} That sounds amazing. I'm more of a couch-and-coffee kind of person, but I get it. Sometimes you just need to move, you know? Whether it's on two wheels or two feet, it's about feeling alive. And hey, if you ever need a running buddy, I'll try not to slow you down too much.

\textbf{Speaker 1:} Deal. But fair warning---I'm not slowing down for anyone. You'll have to keep up. And if you're lucky, I might even let you borrow my helmet. Just don't scratch it.

\textbf{Speaker 2:} Oh, I'll take my chances. But seriously, you're inspiring. I think I'm going to start running again. Maybe not a marathon---baby steps, right? Like, a 5K. Or even just to the corner store.

\textbf{Speaker 1:} That's the spirit! Start small, build up. And hey, if you ever want to ride shotgun on the bike, just say the word. I'll even let you pick the playlist. As long as it's not country.

\textbf{Speaker 3:} Country? No way. I'm more of a punk rock kind of person. But I'll settle for some good indie vibes. And maybe a post-run coffee? I know this place downtown that has the best oat milk lattes.

\textbf{Speaker 1:} Sold. Let's make it a thing. Run, ride, repeat. And who knows? Maybe next year, we'll all be at the marathon together. {\color{red} You in the blue shirt (Image 4)}, {\color{red} me in the white fleece (Image 3)}, and you, Speaker 2, in whatever ridiculous outfit you can find.

\textbf{Speaker 2:} I'm already planning my costume. Think glitter and neon. And maybe a cape. Because if you're going to run a marathon, you might as well do it with flair.

\textbf{Speaker 3:} Glitter and neon? I'm in. Let's make it legendary.

\end{document}